\documentclass[letterpaper]{article} 
\usepackage{aaai2026}  
\usepackage{times}  
\usepackage{helvet}  
\usepackage{courier}  
\usepackage[hyphens]{url}  
\usepackage{graphicx} 
\usepackage{natbib}  
\usepackage{caption} 
\usepackage{amsmath}
\usepackage{amssymb}    
\usepackage{mathtools}  
\usepackage{booktabs}
\usepackage{multirow}
\usepackage{makecell}
\usepackage{amsthm}
\newtheorem{definition}{Definition}
\newtheorem{proposition}{Proposition}
\newtheorem{theorem}{Theorem}

\usepackage{algorithm}
\usepackage{algorithmic}

\nocopyright

\usepackage{newfloat}
\usepackage{listings}
\DeclareCaptionStyle{ruled}{labelfont=normalfont,labelsep=colon,strut=off} 
\floatstyle{ruled}
\newfloat{listing}{tb}{lst}{}
\floatname{listing}{Listing}
\title{SAJA: A State-Action Joint Attack Framework \\ on Multi-Agent Deep Reinforcement Learning}
\author{
    Weiqi Guo\textsuperscript{\rm 1},
    Guanjun Liu\textsuperscript{\rm 1}\thanks{Corresponding author.},
    Ziyuan Zhou\textsuperscript{\rm 1}
}
\affiliations{
    \textsuperscript{\rm 1}Department of Computer Science, Tongji University, Shanghai 201804, China\\
    weiqi\_gwq@tongji.edu.cn, liuguanjun@tongji.edu.cn, ziyuanzhou@tongji.edu.cn
}

\usepackage{bibentry}

\begin{document}

\maketitle

\begin{abstract}
Multi-Agent Deep Reinforcement Learning (MADRL) has shown potential for cooperative and competitive tasks such as autonomous driving and strategic gaming. However, models trained by MADRL are vulnerable to adversarial perturbations on states and actions. Therefore, it is essential to investigate the robustness of MADRL models from an attack perspective. 
Existing studies focus on either state-only attacks or action-only attacks, but do not consider how to effectively joint them. Simply combining state and action perturbations such as randomly perturbing states and actions does not exploit their potential synergistic effects. In this paper, we propose the State-Action Joint Attack (SAJA) framework that has a good synergistic effects. SAJA consists of two important phases: (1) In the state attack phase, a multi-step gradient ascent method utilizes both the actor network and the critic network to compute an adversarial state, and (2) in the action attack phase, based on the perturbed state, a second gradient ascent uses the critic network to craft the final adversarial action. Additionally, a heuristic regularizer measuring the distance between the perturbed actions and the original clean ones is added into the loss function to enhance the effectiveness of the critic's guidance. We evaluate SAJA in the Multi-Agent Particle Environment (MPE), demonstrating that (1) it outperforms and is more stealthy than state-only or action-only attacks, and (2) existing state or action defense methods cannot defend its attacks.
\end{abstract}


\section{Introduction}

Multi-Agent Deep Reinforcement Learning (MADRL) has been successfully applied to challenges such as virtual games \cite{vinyals2019alphastar,berner2019dota} and physical tasks \cite{chen2017decentralized,lin2018efficient,zhang2022rlcharge,chen2023robust}. MADRL methods often employ the actor-critic network with the centralized training and decentralized execution (CTDE) framework, such as MADDPG \cite{lowe2017multi}, COMA \cite{foerster2018counterfactual}, MAPPO \cite{yu2022surprising} and FACMAC \cite{peng2021facmac}. This paradigm is widely adopted for its effectiveness in handling high-dimensional continuous action spaces. However, MADRL algorithms are vulnerable to adversarial attacks \cite{lin2020robustness}, because they are built upon deep neural networks. Studying such attacks is important as it helps to uncover the potential risks of a model and facilitates the generation of perturbations in adversarial training \cite{behzadan2017whatever,tessler2019action}. Therefore, it is necessary to research adversarial attacks.

Existing studies on adversarial attacks mainly focus on state and action dimension. In state attacks, the adversary directly perturbs an agent's input state. Advanced methods often train a dedicated adversary guided by target actions to generate state perturbations. This process is a technique for both direct attacks \cite{lin2020robustness,zhou2023robustness,zhou2024adversarial} and adversarial training \cite{sun2021strongest,han2022solution}. In action attacks, the adversary's goal is to disrupt an agent's behavior, thereby indirectly affecting the next state through environmental interaction. It can be achieved in two ways: either by manipulating an agent's output action \cite{hu2022sparse,sun2022romax}, or by adding adversarial agents into the environment \cite{gleave2019adversarial,li2025attacking}. However, prior work investigates either state attacks or action attacks in isolation, overlooking their potential synergistic effects. 

In this paper we systematically study the state and action joint adversarial attack on multi-agent systems. We design and propose the State-Action Joint Attack (SAJA). SAJA is a novel, efficient, two-phase and gradient-based framework. In the first phase, it uses both the actor and critic network to generate a state perturbation. In the second phase, based on the perturbed state, it again leverages the critic network to craft a subsequent action perturbation. To reduce reliance on the critic network's potentially inaccurate Q-value estimations, a loss function consisting of a heuristic action regularization term and the Q-value is introduced to guide the gradient ascent update process.

Our contributions are as follows:

\begin{itemize}
    \item We formally define the problem of joint state-action adversarial attacks and propose a practical framework (SAJA) to find effective approximate solutions. 
    \item We design a Heuristic Loss Function (HLF), composed of a Q-value term and an action distance term balanced by hyperparameters. Ablation studies demonstrate its superiority over using either term alone, and our theoretical formulation supports the effectiveness of this approach.
    \item We conduct experiments in the Multi-Agent Particle Environment (MPE). Our results demonstrate that SAJA significantly outperforms state-only or action-only attacks in degrading team rewards and can effectively bypass various existing defenses (PAAD, ATLA, M3DDPG). Furthermore, it achieves an equivalent impact with a stealthier perturbation budget than its single-vector counterparts.
\end{itemize}

The remainder of this paper is organized as follows: Section 2 reviews related work, Section 3 introduces the preliminaries, Section 4 details our SAJA method, Section 5 presents and analyzes the experimental results, and Section 6 concludes the paper.

\section{Related Work}

The vulnerability of deep neural networks to adversarial samples has been widely studied. For example, FGSM \cite{goodfellow2014explaining} and PGD \cite{madry2017towards} are widely used as benchmarks for evaluating model robustness. Studies have shown that MADRL policies exhibit vulnerability even to these two methods \cite{huang2017adversarial,tu2021adversarial}. We choose PGD as SAJA's foundational attack technique deliberately. First, to clearly isolate and establish the principle of synergy, we avoid SOTA methods that could confound the analysis. Second, destructive SOTA methods can cause a "saturation effect", making it impossible to fairly measure the additional benefit of a joint attack. Although existing works have proposed many sophisticated state attacks or action attacks (as mentioned in the Introduction), their common limitation is the neglect of the joint attack we focus on. To our knowledge, the only related work, MARLSafe \cite{guo2022towards}, considers multi-dimensional perturbations but fails to explore their interactions.

Researchers have also proposed various adversarial training methods to improve the robustness of MADRL. For state defense, PAAD \cite{sun2021strongest} and ATLA \cite{zhang2021robust} are representative works. PAAD uses a "Director-Actor" mechanism to guide agents to learn to defend against worst-case perturbations in the observation space. ATLA utilizes an alternating training scheme, where an adversary is trained to generate adversarial perturbations. For action defense, M3DDPG \cite{li2019robust} introduces a minimax optimization objective, training agents under the assumption that their counterparts will take adversarial actions. These defense methods are designed for single-dimensional actions or states and may not provide adequate protection against two-dimensional attacks we propose. In this study, we tested whether these defense models can effectively counter joint attack scenarios.

In gradient-based attacks, the design of the loss function largely determines the attack's efficiency and direction. Using only the negative critic's Q-value \cite{li2019robust} as the loss function is unreliable, as imperfect critic training leads to poor gradients. A prior study \cite{zhang2020robust} proposed a heuristic attack method called "Maximal Action Difference" (MAD), but this was never applied to multi-agent or joint attacks. Inspired by MAD, we propose a balanced loss function for SAJA that combines the negative Q-value with an action distance term.

\section{Preliminaries}

In this section, we introduce the standard mathematical formalisms for reinforcement learning and the multi-agent actor-critic framework.

\subsection{Reinforcement Learning Formalisms}

A single-agent reinforcement learning problem is typically modeled as a \textbf{Markov Decision Process (MDP)}, which is defined as $\langle S, A, P, r, \gamma \rangle$, where $S$ is the state space, $A$ is the action space, $P(s'|s, a)$ is the state transition probability function, $r(s, a)$ is the immediate reward function, $\gamma \in [0, 1)$ is the discount factor.

When multiple agents are present in the system, the problem can be modeled as a Markov Game \cite{shapley1953stochastic} or a Decentralized Partially Observable Markov Decision Process\cite{bernstein2002complexity}, depending on the agents' observability.

A \textbf{Markov Game (MG)} assumes that all agents can observe the global state. An $n$-agent MG is defined by $\langle S, \{A_i\}_{i=1}^n, P, \{r_i\}_{i=1}^n, \gamma \rangle$, where $A_i$ is the action space for agent $i$, $P(s'|s, a_1, \dots, a_n)$ is the state transition function which depends on the current state and the joint action of all agents, $r_i(s, a_1, \dots, a_n)$ is the reward function for agent $i$.

A \textbf{Decentralized Partially Observable Markov Decision Process (Dec-POMDP)} assumes that each agent receives only a partial and local observation. This formalism is naturally suited for settings that require decentralized execution. An $n$-agent Dec-POMDP is defined as $\langle S, \{A_i\}_{i=1}^n, P, \{r_i\}_{i=1}^n, \{o_i\}_{i=1}^n, O, \gamma \rangle$, where $o_i$ is the observation space for agent $i$, $O(o_1, \dots, o_n | s, a_1, \dots, a_n)$ is the probability of receiving the joint observation $o$ after taking the joint action $a$ in state $s$.

\subsection{Multi-Agent Actor-Critic}

Actor-Critic (AC) methods are a common approach in MADRL. In this setup, each agent~$i$ uses $\mu_i(o_i)$ to select an action. And critic networks $Q$, access global information to evaluate actors' decisions and guide their updates.

\textbf{MADDPG} is a classic AC algorithm that assigns a separate, centralized critic $Q_i$ to each agent~$i$. The critic network $Q_i(s, a_1, \dots, a_n; \phi_i)$ updates its parameters $\phi_i$ by minimizing the temporal difference (TD) error:
\begin{equation}
\label{eq:maddpg_critic_loss}
\begin{split}
    L(\phi_i) &= E_{(s,a,r,s')\sim\mathcal{D}} \left[ (y_i - Q_i(s, a; \phi_i))^2 \right],\\
    y_i &= r_i + \gamma Q'_{i}(s', a'_1, \dots, a'_n; \phi'_{i}) \big|_{a'_{k}=\mu'_{k}(o'_{k})}
\end{split}
\end{equation}
where $\mathcal{D}$ is the experience replay buffer, $y_i$ is the target value, $Q'_{i}$ is the target critic, $\mu'_{k}$ is the target actor networks and $\gamma$ is the discount factor. 

\textbf{FACMAC} integrates the ideas of MADDPG with value decomposition techniques. FACMAC employs a single, centralized total critic $Q_{\text{tot}}$, shared by all agents. This critic uses a non-linear mixing network $g_{\text{mix}}$ to combine information from all agents.
$Q_{\text{tot}}$ computes the joint-action Q-value by feeding each agent's local Q-value $Q_i(o_i, a_i; \phi_i)$, and the global state $s$ into the mixing network $g_{\text{mix}}$:
\begin{align}
\label{eq:facmac_q_tot}
    Q_{\text{tot}}(s, a; \phi) &= \nonumber \\
    g_{\text{mix}} &(Q_1(o_1,a_1;\phi_1), \dots, Q_n(o_n,a_n;\phi_n), s; \phi_{\text{mix}})
\end{align}
The parameters of the entire critic network $\phi = \{\phi_1, \dots, \phi_n, \phi_{\text{mix}}\}$, are updated by minimizing:
\begin{equation}
\label{eq:facmac_critic_loss}
\begin{split}
    L(\phi) &= {E}_{(s,a,r,s')\sim\mathcal{D}} \left[ (y_{\text{tot}} - Q_{\text{tot}}(s,a; \phi))^2 \right], \\
    y_{\text{tot}} &= r + \gamma Q'_{\text{tot}}(s', a'; \phi')
\end{split}
\end{equation}
This structure allows each actor to implicitly consider the impact of its action on the team's collective interest.

\section{The State-Action Joint Attack Method}

In this section, we first establish the theoretical framework for the joint state-action attack and then provide a practical algorithmic implementation.

\subsection{Theoretical Framework}

\begin{definition}[State–Action–Adversarial Markov Game]
A \emph{State–Action–Adversarial Markov Game} is a tuple
\[
\mathcal{M}_{\mathrm{SA}} = \bigl\langle \mathcal{N}, \;\mathcal{S},\;\{\mathcal{A}_i\}_{i=1}^N,\;P,\;r\bigr\rangle,
\]
\end{definition}

where $\mathcal{N}$ is the agents' set; $S$ is the global state space; ${A}_i$ is the action space of agent $i$; $P(s'\mid s,a_1,\dots,a_N)$ is the state–transition and $r(s,a)$ is the team reward. At each step, an adversary observes the true state $s$ and can introduce perturbations $\delta_s$ and $\delta_a$. 

\begin{definition}[Perturbation Sets]
Let $\|\cdot\|_p$ and $\|\cdot\|_q$ be chosen norms on the state and action spaces. Given radii $\varepsilon_s, \varepsilon_a \ge 0$, the sets of allowed perturbations are defined as
\begin{align}
\label{eq:Perturbation_sets}
    &\Delta_s = \bigl\{\delta_s\in{R}^s : \|\delta_s\|_p \le \varepsilon_s\bigr\},\qquad \\
    &\Delta_{a} = \bigl\{\delta_a\in{R}^{a} : \|\delta_a\|_q \le \varepsilon_a\bigr\}
\end{align}
\end{definition}

In the process each agent computes its action based on $\hat{s} = s + \delta_s$, yielding a joint action $a' = \mu(\hat{s})$. The adversary leads to the final executed action $a'' = a' + \delta_a$. The environment transitions based on the true state $s' \sim P(\cdot \mid s, a'')$. However ,directly solving for a globally optimal attack policy is intractable, we therefore introduce a heuristic transition from global to single-step attack: 
\begin{proposition}[Heuristic Transition]
\label{proposition: heuristic_transition}
    A state-action perturbation that most effectively degrades a high-quality value estimate at a single timestep is highly likely to be a key contributor to the reduction of the final episodic reward.
\end{proposition}

However, this proposition's limitation lies in the myopic nature of the single-step optimization. A globally optimal attacker, for instance, might forgo a large immediate Q-value degradation in favor of a subtle perturbation that lures the agents into a long-term trap state. We now show that under a standard assumption of critic accuracy, the performance gap between the intractable globally optimal attack and our single-step attack is bounded. 

\begin{theorem}[Critic Accuracy]
\label{theorem:critic}
We assume the trained critic network $Q(s,\mathbf{a})$ is an $\epsilon_Q$-approximation of the true team value function $V^{\Pi}(s,\mathbf{a})$ for the joint policy $\Pi$:
\begin{equation}
    |Q(s,\mathbf{a}) - V^{\Pi}(s,\mathbf{a})| \le \epsilon_Q, \quad \forall s \in \mathcal{S}, \mathbf{a} \in \mathcal{A},
\end{equation}
where $V^{\Pi}(s,\mathbf{a}) = r(s,\mathbf{a}) + \gamma \mathbb{E}_{s' \sim P(\cdot|s,\mathbf{a})}[V^{\Pi}(s')]$ is the true value under no attack. Then the gap between the value function of the heuristic single-step attack and the globally optimal attack is bounded.
\end{theorem}

A detailed proof of Theorem~\ref{theorem:critic} is provided in Appendix~\ref{section:appendix_a}. This allows us to simplify the SAA-MG into an effective single-timestep attack.Therefore, our goal is to find an optimal pair of perturbations $(\delta_s^*, \delta_a^*)$ at each timestep . This single-timestep joint optimization problem can be defined as:
\begin{equation}
    \label{eq:single_step_joint}
    (\delta_s^*, \delta_a^*) = \underset{\substack{\delta_s \in \Delta_s \\ \delta_a \in \Delta_a}}{\operatorname{argmin}} Q(s + \delta_s, \mu(s + \delta_s) + \delta_a).
\end{equation}

However, jointly optimizing $\delta_s$ and $\delta_a$ is a high-dimensional, non-convex challenge. We therefore decompose it into a more tractable two-phase sequential optimization problem:

First, we should find an optimal state perturbation $\delta_s^*$ that minimizes the critic's Q-value after the actors take actions based on $s' = s + \delta_s^*$.
\begin{equation}
    \label{eq:seq_state_attack}
    \delta_s^* = \underset{\delta_s \in \Delta_s}{\operatorname{argmin}} Q(s + \delta_s, \mu(s + \delta_s))
\end{equation}
    
Second, based on the perturbed state $s^*=s+\delta_s^*$, we should find an optimal action perturbation $\delta_a^*$ such that the final executed action $a^{**} = \mu(s^*) + \delta_a^*$ further minimizes the critic's Q-value.
\begin{equation}
    \label{eq:seq_action_attack}
    \delta_a^* = \underset{\delta_a \in \Delta_a}{\operatorname{argmin}} Q(s^*, \mu(s^*) + \delta_a)
\end{equation}

This theoretical framework provides a principled guide for practical joint attack implementation. Next, we introduce SAJA, an efficient, gradient-based algorithm for approximately solving this two-stage sequential optimization problem.

\subsection{SAJA Algorithm}

To practically implement the sequential optimization goals, we propose the \textbf{State-Action Joint Attack (SAJA)}, detailed in Algorithm~\ref{alg:saja}. SAJA is a two-phase method, extending the principles of PGD \cite{tu2021adversarial}. SAJA proceeds in two phases within a single timestep: first crafting the state perturbation (State Attack Phase), and then using the perturbed state to craft the final action perturbation (Action Attack Phase).

\paragraph{Victim Selection} At each timestep, SAJA randomly selects a subset of $m$ agents from the total agents $1,2,\dots,n$ to act as victims. The attack perturbations are computed for and applied only to this subset, which allows the attack to be more targeted and potentially less detectable than attacking all agents simultaneously.

\paragraph{Heuristic Loss Function}

In the single-agent setting, \cite{zhang2020robust} proposed Maximal Action Difference (MAD) and demonstrated that an effective adversarial state $\hat{s}$ can be found. We extend this principle to the multi-agent context. First, we define the Multi-Agent Action Difference (MAAD).

\begin{definition}[Multi-Agent Action Difference]
Given a joint policy $\Pi = \{\pi_1, \dots, \pi_n\}$ composed of $n$ individual agent policies, the $\mathcal{D}_{\Pi}(s, \hat{s})$ between a true state $s$ and a perturbed state $\hat{s}$ is defined as the sum of the KL-divergences of all individual policies:
\begin{equation}
    \mathcal{D}_{\Pi}(s, \hat{s}) = \sum_{i=1}^{n} D_{\text{KL}}(\pi_i(\cdot|s) \,||\, \pi_i(\cdot|\hat{s})).
    \label{eq:ma_ad}
\end{equation}
\end{definition}

Based on this definition, we can establish a connection between the action difference and the degradation in team reward.

\begin{theorem}[Bounded Reward Degradation]
\label{theorem:bounded_reward_degradation}
Assuming a bounded reward function, for a given joint policy $\Pi$ under a worst-case adversarial state attack $\hat{s}^* = \arg\max_{\hat{s} \in \mathcal{B}(s)} \mathcal{D}_{\Pi}(s, \hat{s})$, the degradation of the value function is bounded and positively correlated with the upper bound of the multi-agent action difference:
\begin{equation}
    |V^{\Pi}(s) - V^{\Pi}_{\text{adv}}(s)| \le C \cdot \max_{\hat{s} \in \mathcal{B}(s)} \sqrt{\mathcal{D}_{\Pi}(s, \hat{s})},
\end{equation}
where $C$ is a constant and $V^{\Pi}_{\text{adv}}(s)$ is the value under the optimal adversarial attack.
\end{theorem}

Theorem~\ref{theorem:bounded_reward_degradation} provides the key theoretical support for the design of our heuristic loss function. It reveals that maximizing the MAAD is a principled objective for an attacker seeking to degrade the team's value. A detailed proof is provided in Appendix~\ref{section:appendix_a}. 

From Theory to Practice, we introduce a Heuristic Loss Function (HLF) that combines the standard negative $Q$-value objective with a heuristic regularization term. It is defined as the $\ell_2$ distance between the perturbed action and the original clean action. The final HLF for the state-attack phase ($L_s$) and the action-attack phase ($L_a$) are defined as a weighted sum:
\begin{equation}
    L_s \leftarrow -\alpha_1 Q(s_{\text{adv}}^{(k-1)}, a') + \beta_1 \|a' - a^{(0)}\|_2,
    \label{eq:loss_state}
\end{equation}
\begin{equation}
    L_a \leftarrow -\alpha_2 Q(s^*, a_{\text{adv}}^{(l-1)}) + \beta_2 \|a_{\text{adv}}^{(l-1)} - a^{(0)}\|_2.
    \label{eq:loss_action}
\end{equation}
Here, the hyperparameters $\alpha$ and $\beta$ control the balance between deceiving the critic and inducing behavioral deviation.

\begin{algorithm}[t]
\caption{State-Action Joint Attack (SAJA)}
\label{alg:saja}
\textbf{Input}: Original state $s$, actor networks $\mu$, critic network $Q$, perturbation budgets $\varepsilon_s, \varepsilon_a$. \\
\textbf{Parameters}: Attack iterations $K_s, K_a$, step sizes $\alpha_s, \alpha_a$, loss weights $\alpha_1, \beta_1, \alpha_2, \beta_2$. \\
\textbf{Output}: Adversarial state-action pair $(s^*, a^{**})$. \\
\begin{algorithmic}[1] 
\STATE Get original state $s$ 
\STATE Compute original action $a^{(0)} \leftarrow \mu(s)$.
\STATE Randomly select a set $V$ of $m$ victims from $\{1, \dots, n\}$.
\STATE Initialize adversarial state $s_{adv}^{(0)} \leftarrow s$. 

\STATE \textit{// --- State Attack Phase ---}
\FOR{$k = 1$ \TO $K_s$}
    \STATE Compute intermediate action from perturbed state: \\
    $a' \leftarrow \mu(s_{adv}^{(k-1)})$.
    \STATE Define loss using Eq. (\ref{eq:loss_state}).
    \STATE Update $m$ victims' states: \\
    $s_{adv, V}^{(k)} \leftarrow s_{adv, V}^{(k-1)} + \alpha_s \cdot \text{sign}(\nabla_{s_{adv, V}^{(k-1)}} L_s)$.
    \STATE Project onto $\varepsilon_s$-ball: \\
    $s_{adv}^{(k)} \leftarrow s + \text{clip}(s_{adv}^{(k)} - s, -\varepsilon_s, \varepsilon_s)$.
\ENDFOR
\STATE Set final perturbed state $s^* \leftarrow s_{adv}^{(K_s)}$.

\STATE \textit{// --- Action Attack Phase ---}
\STATE Initialize adversarial action $a_{adv}^{(0)} \leftarrow \mu(s^*)$.
\FOR{$l = 1$ \TO $K_a$}
    \STATE Define loss using Eq. (\ref{eq:loss_action}).
    \STATE Update $m$ victims' actions: \\
    $a_{adv, V}^{(l)} \leftarrow a_{adv, V}^{(l-1)} + \alpha_a \cdot \text{sign}(\nabla_{a_{adv, V}^{(l-1)}} L_a)$.
    \STATE Project onto $\varepsilon_a$-ball: \\
    $a_{adv}^{(l)} \leftarrow \mu(s^*) + \text{clip}(a_{adv}^{(l)} - \mu(s^*), -\varepsilon_a, \varepsilon_a)$.
\ENDFOR
\STATE Set final perturbed action $a^{**} \leftarrow a_{adv}^{(K_a)}$.

\STATE \textbf{return} $(s^*, a^{**})$
\end{algorithmic}
\end{algorithm}

\section{Experiments}

\subsection{Experimental Settings}

\paragraph{Environments} 
We conduct our experiments using the open-source code framework provided with the FACMAC paper \cite{peng2021facmac}, which includes the Multi-Agent Particle Environment \cite{lowe2017multi}. We select two continuous predator-prey scenarios: three agents and one prey (3a scenario) , and six agents and two preys (6a scenario). Each agent's observation is a 16-dimensional vector for the 3a scenario and a 30-dimensional vector for the 6a scenario. Actions are 2-dimensional vectors for both.

\paragraph{Benchmark MADRL Models}
Our experiments are based on two actor-critic algorithms: MADDPG and FACMAC. We create two families of target policies for our attacks: vanilla models and defended models.

\begin{itemize}
    \item Vanilla Models: We trained them using the default parameters from the repository, denoted as \textbf{3a-MADDPG}, \textbf{6a-MADDPG}, \textbf{3a-FACMAC}, and \textbf{6a-FACMAC}.
    \item Defended models: For state defenses, we adopt PAAD and ATLA, with implementations following the methodology in \cite{zhou2024adversarial}. For action defense, we employed M3DDPG and modified its implementation to fit our shared-critic framework by jointly perturbing all agents' actions during each update. More details of defense methods are provided in Appendix~\ref{section:appendix_b}. To ensure a meaningful attack evaluation, we only selected models whose test mean rewards were comparable to or higher than the vanilla counterparts. The final set includes: \textbf{3a-PAAD-MADDPG}, \textbf{3a-ATLA-MADDPG}, \textbf{3a-m3ddpg}, \textbf{6a-PAAD-FACMAC}, and \textbf{6a-ATLA-FACMAC}.
\end{itemize}

\paragraph{Benchmark Attack Methods}
To evaluate SAJA, we designed two classes of baselines for a fair comparison, as no prior joint attack methods exist. Our principle is that all gradient-based methods share the same technique (PGD), so we can focus on the "joint" attack mechanism itself. More details are provided in Appendix~\ref{section:appendix_c}.

\begin{itemize}
    \item Stochastic Attack: We use three stochastic attacks to demonstrate that SAJA's effectiveness comes from its gradient-based optimization, not only from its injected noise or its two-phase structure. \textbf{Random-State} and \textbf{Random-Action} apply random sign perturbation to the state and action respectively. And \textbf{Random-State-Action} follows the same framework as SAJA but replaces the gradient-based state and action attacks also with random sign perturbation.
    \item Ablations of SAJA: We compare SAJA against its two phases as ablated baselines: \textbf{PGD-State}, which applies only the state attack phase of SAJA, and \textbf{PGD-Action}, which applies only its action attack phase. This comparison on an identical PGD-based foundation allows for a fair assessment of the joint attack. We do not benchmark SAJA with state-of-the-art (SOTA) single-vector attacks, because their different underlying techniques would confound the joint attack mechanism analysis.
\end{itemize}

\paragraph{Perturbation and Hyperparameter Settings} Except for the budget allocation experiments, we set the state perturbation norm $\epsilon_s$ to 0.02 and the action perturbation norm $\epsilon_a$  to 0.05. All of the gradient-based attacks' optimization steps are set to 20. The weights for the Q-value term $\alpha$ and the action regularization term $\beta$ in the heuristic loss function are set to 0.01 and 0.99, respectively, to balance the two terms. For evaluation, each experimental configuration is run 5 times with different random seeds, with each run conducting 500 episodes. For each run, we compute its average episodic team reward and its variance. We finally report the mean of the five average rewards and the mean of the five variances. A lower mean reward indicates a more effective attack. More details of experimental settings are provided in Appendix~\ref{section:appendix_d}.

\subsection{Results Analysis}

We evaluate the effectiveness of SAJA on both vanilla and defended models, and compare it against five benchmark attack methods. The results are summarized in Table \ref{tab:vanilla_effectiveness} and Table \ref{tab:defense_effectiveness}. In each table, we report the average episodic team reward for various target models and numbers of attacked agents. Each reported value represents the mean and standard deviation over 5 runs. The compared methods include three random attacks (R-State for Random-State, R-Action for Random-Action, and R-SA for Random-State-Action), and three gradient-based attacks (PGD-S for PGD-State, PGD-A for PGD-Action, and SAJA). The two tables also include the reward under no attack and the percentage reward drop caused by SAJA.

In Table \ref{tab:vanilla_effectiveness}, our analysis proceeds on three levels. First, compared to the no-attack baseline, both Random-State and Random-Action induce a slight reward degradation. This confirms the multi-agent models' general sensitivity to noise. Second, we compare the effects of single-vector versus joint attacks. Among random attacks, Random-State-Action achieves a lower reward than Random-State and Random-Action in 6 out of 12 settings. This phenomenon becomes more apparent in SAJA, which outperforms both PGD-State and PGD-Action in 10 out of 12 settings. This trend strongly suggests that the more directed a perturbation becomes, shifting from random noise to objective-driven methods, the more effectively a joint attack can exploit the synergistic vulnerabilities between continuous state and action spaces. This finding provides a powerful validation of our SAJA core design principle. Finally, from a model-specific perspective, SAJA is most effective against the 3a-FACMAC, causing the Reward Drop up to $9.90\%$. However, all attacks including SAJA, are largely ineffective against the 6a-MADDPG, whose low baseline reward offers no meaningful gradient for exploitation. This indicates that gradient-based attacks can only rely on a well-trained target model.

For the analysis of attacking defended models, we first clarify that although M3DDPG is categorized as an action defense, its mechanism indirectly influences an agent's state by changing other agents' actions. Its goal thus is similar to direct state defenses like PAAD and ATLA. Therefore, we treat them as the same mechanism designed to enhance model robustness. In Table \ref{tab:defense_effectiveness}, we find that the quality of adversarial training directly influences SAJA's efficiency. On 3a-PAAD-MADDPG, 3a-ATLA-MADDPG, and 3a-M3DDPG, the Random-State attack causes only a slight degradation in reward, indicating that their defense methods effectively resist random sign perturbation. However, this simultaneously enhances the quality of the actor and critic gradients in the face of perturbations, allowing SAJA to exploit a more potent joint attack. Evidence for this is that the SAJA Reward Drop for 3a-PAAD-MADDPG $(4.01\%,6.92\%,6.89\%)$, 3a-ATLA-MADDPG $(5.06\%,4.69\%,6.96\%)$ and 3a-M3DDPG $(3.85\%,6.27\%,6.55\%)$ is significantly higher than that for 3a-MADDPG $(2.64\%,4.52\%,5.44\%)$. Conversely, on 6a-PAAD-FACMAC and 6a-ATLA-FACMAC, the Random-State attack causes a significant reward drop, suggesting inadequate  adversarial training. The SAJA Reward Drop for 6a-PAAD-FACMAC and 6a-ATLA-FACMAC is less than that for 6a-FACMAC. 

In summary, we conclude that: (1) Joint attacks are more effective than their corresponding single-vector attacks. This synergy is particularly pronounced in gradient-based methods. (2) Current gradient-based methods like SAJA have only begun to exploit the surface of this state-action synergy. It suggests significant potential for developing more potent joint attack algorithms. (3) A better-trained model, such as a robustly defended one, may become more vulnerable to SAJA. This is because its higher-quality training provides more precise gradients, which our attack method directly uses to craft more devastating perturbations.

\begin{table*}[t]
\centering
\small
\setlength{\tabcolsep}{0.5mm} 
\begin{tabular}{l|c|cccccc|c}
\hline
\multirow{2}{*}{\textbf{\makecell{Vanilla Models \\ (Base Reward)}}} & \multirow{2}{*}{\textbf{Vic.}} & \multicolumn{6}{c|}{\textbf{Attack Method}} & \multirow{2}{*}{\textbf{\makecell{SAJA \\ Drop}}} \\
\cline{3-8}
& & \textbf{R-State} & \textbf{R-Action} & \textbf{R-SA} & \textbf{PGD-S} & \textbf{PGD-A} & \textbf{SAJA (Ours)} & \\
\hline
\multirow{3}{*}{\makecell{3a-FACMAC \\ 194.66 $\pm$ 95.62}} & 1 & 186.11 $\pm$ 92.59 & 192.99 $\pm$ 95.37 & 187.50 $\pm$ 92.52 & 189.35 $\pm$ 93.15 & 190.51 $\pm$ 96.49 & \textbf{184.63 $\pm$ 91.99} & 5.15\% \\
& 2 & 183.43 $\pm$ 89.76 & 194.90 $\pm$ 95.07 & 183.03 $\pm$ 90.36 & 185.41 $\pm$ 90.70 & 187.67 $\pm$ 93.33 & \textbf{179.62 $\pm$ 88.80} & 7.73\% \\
& 3 & 180.58 $\pm$ 90.03 & 190.99 $\pm$ 94.43 & 177.22 $\pm$ 89.28 & 189.20 $\pm$ 90.88 & 179.34 $\pm$ 92.03 & \textbf{175.38 $\pm$ 90.22} & 9.90\% \\
\hline
\multirow{3}{*}{\makecell{3a-MADDPG \\ 134.59 $\pm$ 70.46}} & 1 & 130.63 $\pm$ 70.09 & 132.90 $\pm$ 69.11 & 130.76 $\pm$ 68.30 & 133.50 $\pm$ 69.40 & \textbf{130.08 $\pm$ 68.56} & 131.04 $\pm$ 68.03 & 2.64\% \\
& 2 & 131.14 $\pm$ 67.01 & 133.24 $\pm$ 68.49 & 129.97 $\pm$ 67.53 & 132.87 $\pm$ 69.84 & 128.85 $\pm$ 68.29 & \textbf{128.50 $\pm$ 69.11} & 4.52\% \\
& 3 & 129.09 $\pm$ 67.24 & 133.09 $\pm$ 68.56 & 128.08 $\pm$ 66.97 & 131.78 $\pm$ 71.91 & 129.74 $\pm$ 68.08 & \textbf{127.26 $\pm$ 69.65} & 5.44\% \\
\hline
\multirow{3}{*}{\makecell{6a-FACMAC \\ 311.55 $\pm$ 105.51}} & 1 & 307.74 $\pm$ 101.85 & 312.61 $\pm$ 105.22 & \textbf{305.50 $\pm$ 104.33} & 309.75 $\pm$ 106.78 & 307.75 $\pm$ 103.16 & 307.44 $\pm$ 103.20 & 1.32\% \\
& 2 & 304.94 $\pm$ 102.02 & 311.09 $\pm$ 104.74 & 306.79 $\pm$ 100.64 & 307.61 $\pm$ 103.43 & 306.29 $\pm$ 101.68 & \textbf{302.28 $\pm$ 101.93} & 2.97\% \\
& 3 & 306.83 $\pm$ 98.68 & 312.33 $\pm$ 103.70 & 304.42 $\pm$ 100.32 & 306.98 $\pm$ 103.24 & 301.19 $\pm$ 103.38 & \textbf{296.57 $\pm$ 99.75} & 4.81\% \\
\hline
\multirow{3}{*}{\makecell{6a-MADDPG \\ 33.67 $\pm$ 30.93}} & 1 & 33.01 $\pm$ 30.19 & 33.84 $\pm$ 30.95 & 33.14 $\pm$ 31.05 & 33.72 $\pm$ 30.59 & 32.98 $\pm$ 30.76 & \textbf{32.85 $\pm$ 29.86} & 2.43\% \\
& 2 & 33.34 $\pm$ 30.38 & 33.88 $\pm$ 31.25 & 33.62 $\pm$ 30.84 & 33.98 $\pm$ 30.90 & \textbf{32.79 $\pm$ 30.01} & 33.12 $\pm$ 30.92 & 1.63\% \\
& 3 & \textbf{33.02 $\pm$ 30.21} & 33.86 $\pm$ 30.50 & 33.98 $\pm$ 30.37 & 33.78 $\pm$ 31.16 & 34.44 $\pm$ 31.60 & 33.50 $\pm$ 30.57 & 0.50\% \\
\hline
\end{tabular}
\caption{Performance comparison against vanilla models. The rows represent the vanilla models and the number of attacked agents (Vic.). The columns list the average episodic team reward (mean $\pm$ std) under various attack methods. Bold values highlight the best attack performance in each row. The final column shows the percentage reward drop caused by SAJA relative to the no-attack baseline (Base Reward). The perturbation budgets are fixed at $\epsilon_s=0.02$ and $\epsilon_a=0.05$.}
\label{tab:vanilla_effectiveness}
\end{table*}

\begin{table*}[t]
\centering
\small
\setlength{\tabcolsep}{0.4mm} 
\begin{tabular}{l|c|cccccc|c}
\hline
\multirow{2}{*}{\textbf{\makecell{Defended Models \\ (Base Reward)}}} & \multirow{2}{*}{\textbf{Vic.}} & \multicolumn{6}{c|}{\textbf{Attack Method}} & \multirow{2}{*}{\textbf{\makecell{SAJA \\ Drop}}} \\
\cline{3-8}
& & \textbf{R-State} & \textbf{R-Action} & \textbf{R-SA} & \textbf{PGD-S} & \textbf{PGD-A} & \textbf{SAJA (Ours)} & \\
\hline
\multirow{3}{*}{\makecell{3a-PAAD-MADDPG \\ 140.36 $\pm$ 73.09}} & 1 & 140.65 $\pm$ 72.05 & 138.20 $\pm$ 72.94 & 137.35 $\pm$ 72.16 & 139.21 $\pm$ 72.62 & 137.72 $\pm$ 73.25 & \textbf{134.73 $\pm$ 73.67} & 4.01\% \\
& 2 & 136.30 $\pm$ 73.74 & 137.78 $\pm$ 72.63 & 137.47 $\pm$ 72.39 & 137.64 $\pm$ 75.86 & 133.81 $\pm$ 72.93 & \textbf{130.64 $\pm$ 73.85} & 6.92\% \\
& 3 & 137.09 $\pm$ 71.10 & 139.90 $\pm$ 73.37 & 135.53 $\pm$ 72.15 & 135.67 $\pm$ 76.63 & 133.05 $\pm$ 72.00 & \textbf{130.69 $\pm$ 73.95} & 6.89\% \\
\hline
\multirow{3}{*}{\makecell{3a-ATLA-MADDPG \\ 122.34 $\pm$ 71.30}} & 1 & 123.27 $\pm$ 70.57 & 122.72 $\pm$ 69.84 & 120.34 $\pm$ 71.38 & 121.58 $\pm$ 71.30 & 120.71 $\pm$ 70.97 & \textbf{116.15 $\pm$ 68.62} & 5.06\% \\
& 2 & 118.91 $\pm$ 69.17 & 119.06 $\pm$ 70.45 & 119.50 $\pm$ 69.40 & 119.99 $\pm$ 70.04 & \textbf{116.22 $\pm$ 69.76} & 116.60 $\pm$ 68.88 & 4.69\% \\
& 3 & 121.06 $\pm$ 69.81 & 122.21 $\pm$ 68.80 & 120.07 $\pm$ 68.79 & 119.83 $\pm$ 70.74 & 114.99 $\pm$ 68.36 & \textbf{113.83 $\pm$ 68.68} & 6.96\% \\
\hline
\multirow{3}{*}{\makecell{3a-M3DDPG \\ 136.30 $\pm$ 70.84}} & 1 & 133.76 $\pm$ 69.36 & 138.81 $\pm$ 71.56 & 132.79 $\pm$ 70.93 & 135.27 $\pm$ 69.84 & 132.94 $\pm$ 69.68 & \textbf{131.06 $\pm$ 69.67} & 3.85\% \\
& 2 & 133.15 $\pm$ 68.93 & 134.04 $\pm$ 70.55 & 131.56 $\pm$ 67.39 & 135.27 $\pm$ 69.84 & 133.14 $\pm$ 70.71 & \textbf{127.76 $\pm$ 67.49} & 6.27\% \\
& 3 & 132.10 $\pm$ 67.69 & 136.89 $\pm$ 72.51 & 129.78 $\pm$ 67.42 & 136.52 $\pm$ 71.27 & 129.45 $\pm$ 70.73 & \textbf{127.37 $\pm$ 67.97} & 6.55\% \\
\hline
\multirow{3}{*}{\makecell{6a-PAAD-FACMAC \\ 314.36 $\pm$ 102.81}} & 1 & \textbf{305.44 $\pm$ 103.38} & 313.42 $\pm$ 102.71 & 312.47 $\pm$ 101.31 & 313.69 $\pm$ 103.15 & 312.23 $\pm$ 101.32 & 308.76 $\pm$ 102.55 & 1.78\% \\
& 2 & 304.99 $\pm$ 100.40 & 315.22 $\pm$ 104.03 & 308.22 $\pm$ 101.50 & \textbf{303.58 $\pm$ 100.25} & 309.88 $\pm$ 102.42 & 307.51 $\pm$ 102.12 & 2.18\% \\
& 3 & 305.97 $\pm$ 99.65 & 316.77 $\pm$ 102.48 & 299.58 $\pm$ 97.50 & 306.91 $\pm$ 101.95 & 307.88 $\pm$ 100.89 & \textbf{299.25 $\pm$ 100.12} & 4.81\% \\
\hline
\multirow{3}{*}{\makecell{6a-ATLA-FACMAC \\ 316.32 $\pm$ 101.58}} & 1 & 313.68 $\pm$ 104.73 & 319.66 $\pm$ 107.19 & 312.74 $\pm$ 103.96 & 314.91 $\pm$ 104.20 & 312.38 $\pm$ 106.51 & \textbf{310.71 $\pm$ 103.82} & 1.77\% \\
& 2 & 311.35 $\pm$ 102.19 & 317.61 $\pm$ 105.64 & 309.05 $\pm$ 104.39 & 315.93 $\pm$ 104.80 & 313.34 $\pm$ 105.76 & \textbf{307.78 $\pm$ 101.09} & 2.70\% \\
& 3 & 306.57 $\pm$ 99.50 & 317.55 $\pm$ 102.19 & 311.09 $\pm$ 100.59 & 313.40 $\pm$ 102.92 & 311.63 $\pm$ 101.46 & \textbf{306.21 $\pm$ 101.68} & 3.19\% \\
\hline
\end{tabular}
\caption{Performance comparison against defended models. The experimental setup is identical to that in Table \ref{tab:vanilla_effectiveness}.}
\label{tab:defense_effectiveness}
\end{table*}

\subsection{Analysis of SAJA Stealthiness}
\begin{figure}[t]
\centering
\includegraphics[width=1\columnwidth]{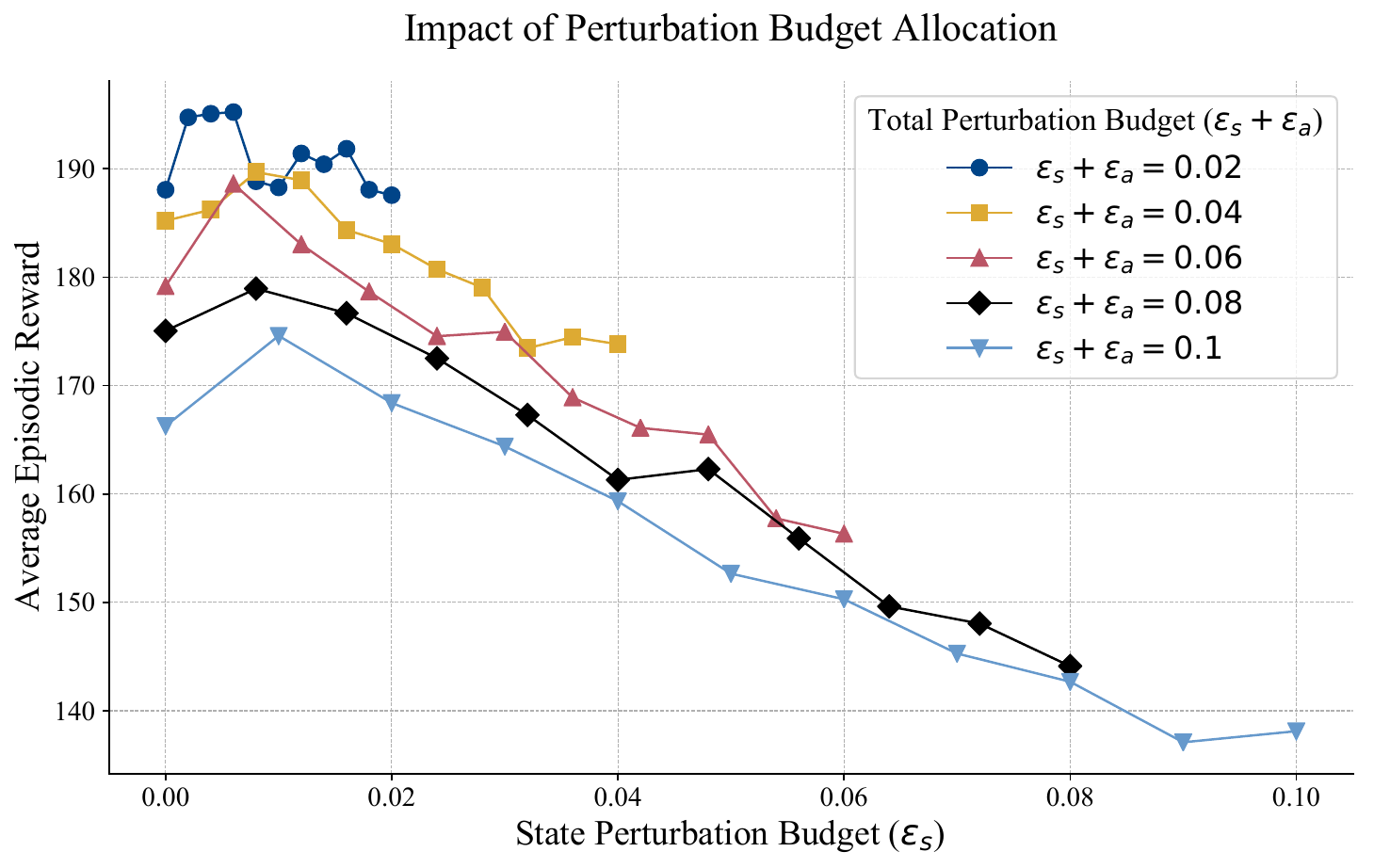}
\caption{Impact of Perturbation Budget Allocation on SAJA. The x-axis represents the state perturbation budget $\epsilon_s$. Each line corresponds to a fixed total budget $s$, as indicated in the legend. The y-axis represents the average episodic reward, with each point calculated as the mean of 5 independent runs (500 episodes each) for each $(\epsilon_s,\epsilon_a)$ setting.}
\label{fig1}
\end{figure}

To evaluate the unique "joint" advantages of SAJA, we designed an experiment to analyze the impact of different perturbation budget allocations. We selected the well-trained 3a-FACMAC as our target model and all three agents as victims. This setting can isolate the effects of perturbation allocation from confounding factors related to model training quality. We defined the total perturbation budget as the sum of the state budget and the action budget $(\epsilon_s+\epsilon_a)$. We set five distinct levels: $S=\{0.02, 0.04, 0.06, 0.08, 0.10\}$. For each total budget $\epsilon_{sa} \in S$, we divided it into 11 different $(\epsilon_s,\epsilon_a)$ combinations, ranging from $(0,\epsilon_{sa})$ (PGD-Action) to $(\epsilon_{sa},0)$ (PGD-State) with a step size of $0.1\epsilon_{sa}$. For example, when $\epsilon_{sa}=0.10$, we tested combinations $(0, 0.10), (0.01, 0.09),\dots, (0.09, 0.01),(0.10,0)$. This resulted in 55 unique budget allocation settings. The performance of SAJA under each setting is plotted in Figure \ref{fig1}. 

\paragraph{Comparison with PGD-Action} As shown in the figure, SAJA $(\epsilon_s=0.03,\epsilon_a=0.03,\epsilon_{sa}=0.06)$ achieves a similar impact to the PGD-Action $(\epsilon_s=0,\epsilon_a=0.08, \epsilon_{sa}=0.08)$. Likewise, the effect of SAJA $\epsilon_s=0.048, \epsilon_a=0.012,\epsilon_{sa}=0.06)$ is comparable to PGD-Action $\epsilon_s=0, \epsilon_a=0.10,\epsilon_{sa}=0.10)$. This shows that SAJA can decompose a large, easily detectable action perturbation into two smaller, less perceptible state and action perturbations, enhancing its stealthiness without sacrificing its effectiveness.

\paragraph{Comparison with PGD-State} In the MPE environment, no instance of SAJA with a smaller $\epsilon_{sa}$ outperforms PGD-State with a larger one. We infer that the MPE-based model is inherently more sensitive to state perturbations. Nevertheless, the advantage of stealthiness persists. For example, the impact of SAJA $(\epsilon_s=0.03,\epsilon_a=0.03,\epsilon_{sa}=0.06)$ is close to PGD-State $(\epsilon_s=0.04,\epsilon_a=0,\epsilon_{sa}=0.04)$. Although the former $\epsilon_{sa}$ is larger, its maximum per-dimension perturbation $(0.03)$ is smaller than that of PGD-State $(0.04)$. Detecting an attack with a perturbation of $(0.03)$ is more difficult than detecting one with a perturbation of $(0.04)$.

We conclude that by distributing the perturbation budget across two dimensions, SAJA can achieve a powerful impact with smaller per-dimension perturbation. It's stealthier than its single-vector counterparts and harder to detect by existing defense mechanisms.

\subsection{Ablation Studies}
\begin{table*}[t]
\centering
\small
\begin{tabular}{l|c|ccc}
\hline
\textbf{Defended Models} & \textbf{Vic.} & \textbf{Q-value Only} & \textbf{Action-Term Only} & \textbf{Balanced (Ours)} \\
\hline
\multirow{3}{*}{3a-PAAD-MADDPG} 
& 1 & 137.43 $\pm$ 74.33 & 137.74 $\pm$ 73.25 & \textbf{134.73 $\pm$ 73.67} \\
& 2 & 136.81 $\pm$ 70.92 & 133.28 $\pm$ 72.06 & \textbf{130.64 $\pm$ 73.85} \\
& 3 & 136.60 $\pm$ 71.86 & \textbf{130.49 $\pm$ 72.10} & 130.69 $\pm$ 73.95 \\
\hline
\multirow{3}{*}{3a-ATLA-MADDPG} 
& 1 & 120.50 $\pm$ 69.41 & 118.89 $\pm$ 70.95 & \textbf{116.15 $\pm$ 68.62} \\
& 2 & 119.64 $\pm$ 68.60 & 118.24 $\pm$ 70.36 & \textbf{116.60 $\pm$ 68.88} \\
& 3 & 113.96 $\pm$ 68.53 & 116.59 $\pm$ 68.05 & \textbf{113.83 $\pm$ 68.68} \\
\hline
\multirow{3}{*}{3a-M3DDPG-MADDPG} 
& 1 & 132.50 $\pm$ 70.03 & 131.07 $\pm$ 69.44 & \textbf{131.06 $\pm$ 69.67} \\
& 2 & 133.91 $\pm$ 68.86 & 128.97 $\pm$ 69.35 & \textbf{127.76 $\pm$ 67.49} \\
& 3 & 129.45 $\pm$ 66.16 & 129.25 $\pm$ 70.35 & \textbf{127.37 $\pm$ 67.97} \\
\hline
\multirow{3}{*}{6a-PAAD-FACMAC} 
& 1 & 311.00 $\pm$ 102.78 & 313.59 $\pm$ 99.24 & \textbf{308.76 $\pm$ 102.55} \\
& 2 & 309.20 $\pm$ 101.86 & \textbf{306.90 $\pm$ 99.66} & 307.51 $\pm$ 102.12 \\
& 3 & 304.16 $\pm$ 100.72 & 303.68 $\pm$ 102.00 & \textbf{299.25 $\pm$ 100.12} \\
\hline
\multirow{3}{*}{6a-ATLA-FACMAC} 
& 1 & 315.46 $\pm$ 101.51 & 316.22 $\pm$ 104.79 & \textbf{310.71 $\pm$ 103.82} \\
& 2 & 311.49 $\pm$ 102.54 & 307.83 $\pm$ 102.89 & \textbf{307.78 $\pm$ 101.09} \\
& 3 & \textbf{304.51 $\pm$ 104.89} & 306.62 $\pm$ 101.59 & 306.21 $\pm$ 101.68 \\
\hline
\end{tabular}%
\caption{Ablation study on the heuristic loss function. The rows represent the defended models and the number of attacked agents (Victims). The columns list the average episodic team reward (mean $\pm$ std) under three different loss function: "Q-value Only" (relying on the critic's value), "Action-Term Only" (relying on the action distance term), and our default "Balanced" method. Bold values highlight the best attack performance in each row.}
\label{tab:loss_ablation}
\end{table*}

\paragraph{State and Action Modules of SAJA}
In Table \ref{tab:vanilla_effectiveness} and Table \ref{tab:defense_effectiveness}, SAJA achieves a lower team reward than both PGD-State (state attack module only) and PGD-Action (action attack module only) in the most cases. This result clearly demonstrates a positive synergy between the two components, showing that their joint is essential for SAJA's superior performance.

\paragraph{Effectiveness of the Heuristic Loss Function}
SAJA utilizes HLF in each of its two phases: $L_s$ (with hyperparameters $\alpha_1,\beta_1$) and $L_a$ (with $\alpha_2,\beta_2$). We use a shared hyperparameter setting by default, where $\alpha_1=\alpha_2=\alpha$ and $\beta_1=\beta_2=\beta$. We designed our ablation study around three settings for the $(\alpha,\beta)$ pair. \textbf{Q-value Only:} $(\alpha,\beta)=(1 - 10^{-6}, 10^{-6})$, representing a traditional attack that relies only on the critic's gradient. \textbf{Action-Term Only:} $(\alpha,\beta)=(10^{-6}, 1-10^{-6})$, representing an attack guided only on our proposed heuristic action distance term. \textbf{Balanced:} $(\alpha,\beta)=(0.01, 0.99)$, which is our default configuration. The results in Table~\ref{tab:loss_ablation} show that the \textbf{Balanced HLF} achieved the lowest reward in 12 out of the 15 configurations. This finding demonstrates that our HLF effectively unites value-based guidance with behavior-based exploration. It not only uncovers deeper state-action synergies but also reduces SAJA's reliance on potentially inaccurate Q-value estimates, thereby enhancing the attack's overall effectiveness.

\subsection{Attack Visualization}

\begin{figure}[t]
\centering
\includegraphics[width=1\columnwidth]{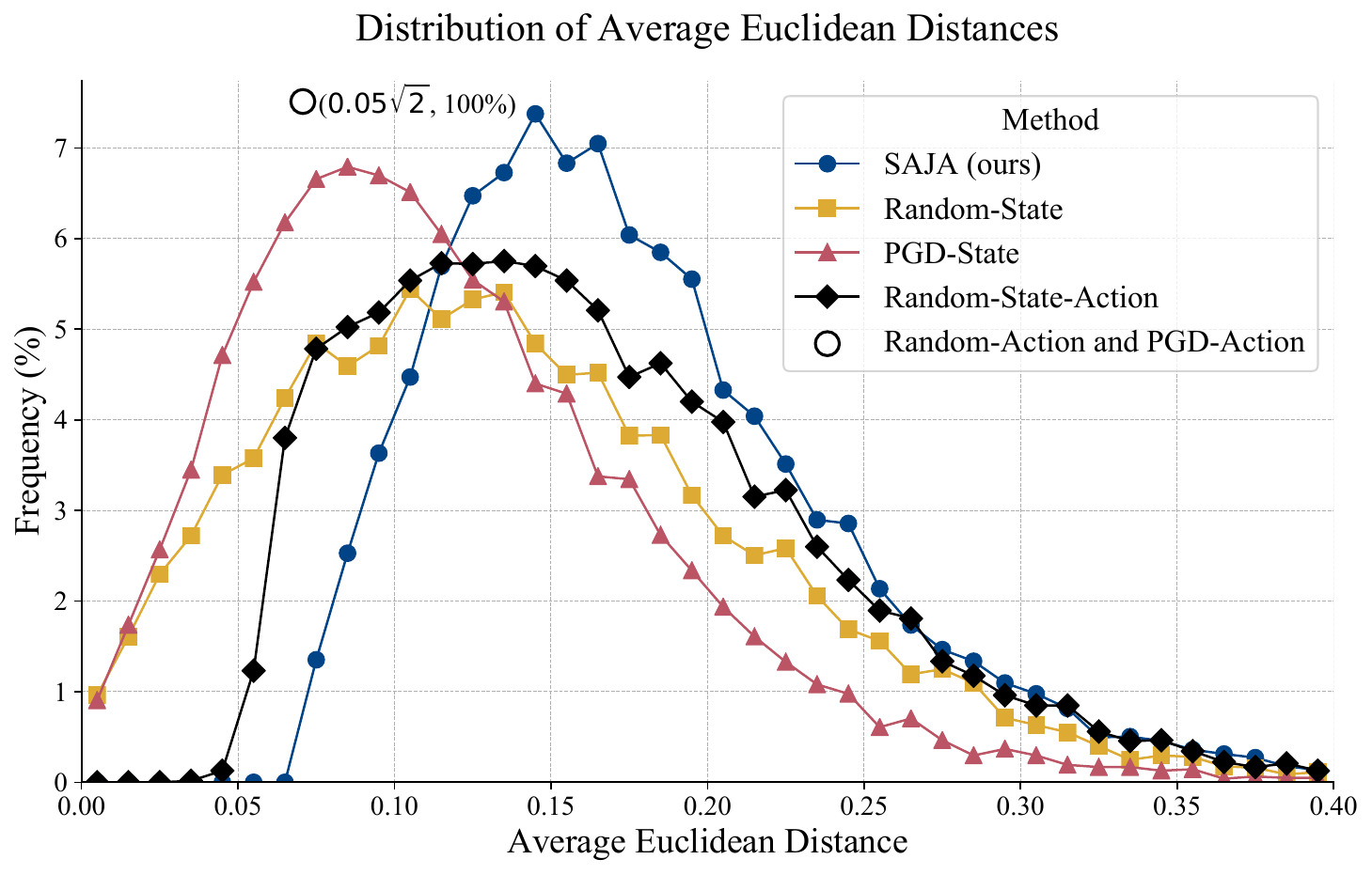}
\caption{ Distribution of the average Euclidean distance under 6 attack methods. The x-axis represents the average L2 distance $||a'' - a^{(0)}||_2 / m$ between the original and final actions of victim agents in an episode. The y-axis shows the frequency of these distances within bins of width 0.01. For clarity, distances greater than 0.40 are omitted due to their low frequency. }
\label{fig2}
\end{figure}

To further investigate the mechanisms of different attack methods, we visualize the degree of action difference they induce. As shown in Figure~\ref{fig2}, we plotted the frequency distribution of the average Euclidean distance, $||a'' - a^{(0)}||_2 / m$, between the final executed actions and the original clean actions over 12,500 timesteps on the 3a-FACMAC. In this setting, the peaks of the action difference distributions for the four methods involving a state-attack phase follow the order PGD-State $<$ Random-State $<$ Random-State-Action $<$ SAJA, which is the exact inverse of their corresponding performance ranking in Table~\ref{tab:vanilla_effectiveness}. This visual evidence verifies our theoretical hypothesis: for well-trained models, maximizing action difference is an effective heuristic attack, thus validating our HLF design.

\section{Conclusion}
In this paper, we addressed a research gap in MADRL system where state and action attacks are not considered. We designed a novel, efficient, two-phase, gradient-based state-action joint attack (SAJA) framework. To reduce the reliance on the critic network's potentially inaccurate Q-value estimations, we innovatively proposed a loss function based on both the Q-value and an action regularization term. Through the theoretical support, and extensive experiments in the MPE environment, we have verified SAJA's superiority and stealthiness against vanilla and defended models.

Nevertheless, as SAJA is a white-box method, a future direction is to explore joint attacks in black-box settings. We also aim to extend SAJA to more complex environments such as StarCraft II \cite{whiteson2019starcraft}. Ultimately, our work may motivate the research of joint defense strategies to counter such sophisticated threats.

\bibliography{aaai2026}

\begin{thebibliography}{32}
\providecommand{\natexlab}[1]{#1}

\bibitem[{Behzadan and Munir(2017)}]{behzadan2017whatever}
Behzadan, V.; and Munir, A. 2017.
\newblock Whatever does not kill deep reinforcement learning, makes it stronger.
\newblock \emph{arXiv preprint arXiv:1712.09344}.

\bibitem[{Berner et~al.(2019)Berner, Brockman, Chan, Cheung, Debiak, Dennison, Farhi, Fischer, Hashme, Hesse et~al.}]{berner2019dota}
Berner, C.; Brockman, G.; Chan, B.; Cheung, V.; Debiak, P.; Dennison, C.; Farhi, D.; Fischer, Q.; Hashme, S.; Hesse, C.; et~al. 2019.
\newblock Dota 2 with large scale deep reinforcement learning.
\newblock \emph{arXiv preprint arXiv:1912.06680}.

\bibitem[{Bernstein et~al.(2002)Bernstein, Givan, Immerman, and Zilberstein}]{bernstein2002complexity}
Bernstein, D.~S.; Givan, R.; Immerman, N.; and Zilberstein, S. 2002.
\newblock The complexity of decentralized control of Markov decision processes.
\newblock \emph{Mathematics of operations research}, 27(4): 819--840.

\bibitem[{Chen et~al.(2023)Chen, Liu, Zhou, Zhang, and Wang}]{chen2023robust}
Chen, S.; Liu, G.; Zhou, Z.; Zhang, K.; and Wang, J. 2023.
\newblock Robust multi-agent reinforcement learning method based on adversarial domain randomization for real-world dual-uav cooperation.
\newblock \emph{IEEE Transactions on Intelligent Vehicles}, 9(1): 1615--1627.

\bibitem[{Chen et~al.(2017)Chen, Liu, Everett, and How}]{chen2017decentralized}
Chen, Y.~F.; Liu, M.; Everett, M.; and How, J.~P. 2017.
\newblock Decentralized non-communicating multiagent collision avoidance with deep reinforcement learning.
\newblock In \emph{2017 IEEE international conference on robotics and automation (ICRA)}, 285--292. IEEE.

\bibitem[{Foerster et~al.(2018)Foerster, Farquhar, Afouras, Nardelli, and Whiteson}]{foerster2018counterfactual}
Foerster, J.; Farquhar, G.; Afouras, T.; Nardelli, N.; and Whiteson, S. 2018.
\newblock Counterfactual multi-agent policy gradients.
\newblock In \emph{Proceedings of the AAAI conference on artificial intelligence}, volume~32.

\bibitem[{Gleave et~al.(2019)Gleave, Dennis, Wild, Kant, Levine, and Russell}]{gleave2019adversarial}
Gleave, A.; Dennis, M.; Wild, C.; Kant, N.; Levine, S.; and Russell, S. 2019.
\newblock Adversarial policies: Attacking deep reinforcement learning.
\newblock \emph{arXiv preprint arXiv:1905.10615}.

\bibitem[{Goodfellow, Shlens, and Szegedy(2014)}]{goodfellow2014explaining}
Goodfellow, I.~J.; Shlens, J.; and Szegedy, C. 2014.
\newblock Explaining and harnessing adversarial examples.
\newblock \emph{arXiv preprint arXiv:1412.6572}.

\bibitem[{Guo et~al.(2022)Guo, Chen, Hao, Yin, Yu, and Li}]{guo2022towards}
Guo, J.; Chen, Y.; Hao, Y.; Yin, Z.; Yu, Y.; and Li, S. 2022.
\newblock Towards comprehensive testing on the robustness of cooperative multi-agent reinforcement learning.
\newblock In \emph{Proceedings of the IEEE/CVF conference on computer vision and pattern recognition}, 115--122.

\bibitem[{Han et~al.(2022)Han, Su, He, Han, Yang, Zou, and Miao}]{han2022solution}
Han, S.; Su, S.; He, S.; Han, S.; Yang, H.; Zou, S.; and Miao, F. 2022.
\newblock What is the solution for state-adversarial multi-agent reinforcement learning?
\newblock \emph{arXiv preprint arXiv:2212.02705}.

\bibitem[{Hu and Zhang(2022)}]{hu2022sparse}
Hu, Y.; and Zhang, Z. 2022.
\newblock Sparse adversarial attack in multi-agent reinforcement learning.
\newblock \emph{arXiv preprint arXiv:2205.09362}.

\bibitem[{Huang et~al.(2017)Huang, Papernot, Goodfellow, Duan, and Abbeel}]{huang2017adversarial}
Huang, S.; Papernot, N.; Goodfellow, I.; Duan, Y.; and Abbeel, P. 2017.
\newblock Adversarial attacks on neural network policies.
\newblock \emph{arXiv preprint arXiv:1702.02284}.

\bibitem[{Li et~al.(2025)Li, Guo, Xiu, Zheng, Feng, Yu, Wang, Liu, Yang, An et~al.}]{li2025attacking}
Li, S.; Guo, J.; Xiu, J.; Zheng, Y.; Feng, P.; Yu, X.; Wang, J.; Liu, A.; Yang, Y.; An, B.; et~al. 2025.
\newblock Attacking cooperative multi-agent reinforcement learning by adversarial minority influence.
\newblock \emph{Neural Networks}, 107747.

\bibitem[{Li et~al.(2019)Li, Wu, Cui, Dong, Fang, and Russell}]{li2019robust}
Li, S.; Wu, Y.; Cui, X.; Dong, H.; Fang, F.; and Russell, S. 2019.
\newblock Robust multi-agent reinforcement learning via minimax deep deterministic policy gradient.
\newblock In \emph{Proceedings of the AAAI conference on artificial intelligence}, volume~33, 4213--4220.

\bibitem[{Lin et~al.(2020)Lin, Dzeparoska, Zhang, Leon-Garcia, and Papernot}]{lin2020robustness}
Lin, J.; Dzeparoska, K.; Zhang, S.~Q.; Leon-Garcia, A.; and Papernot, N. 2020.
\newblock On the robustness of cooperative multi-agent reinforcement learning.
\newblock In \emph{2020 IEEE Security and Privacy Workshops (SPW)}, 62--68. IEEE.

\bibitem[{Lin et~al.(2018)Lin, Zhao, Xu, and Zhou}]{lin2018efficient}
Lin, K.; Zhao, R.; Xu, Z.; and Zhou, J. 2018.
\newblock Efficient large-scale fleet management via multi-agent deep reinforcement learning.
\newblock In \emph{Proceedings of the 24th ACM SIGKDD international conference on knowledge discovery \& data mining}, 1774--1783.

\bibitem[{Lowe et~al.(2017)Lowe, Tamar, Harb, Pieter~Abbeel, and Mordatch}]{lowe2017multi}
Lowe, R.; Tamar, A.; Harb, J.; Pieter~Abbeel, O.; and Mordatch, I. 2017.
\newblock Multi-agent actor-critic for mixed cooperative-competitive environments.
\newblock \emph{Advances in neural information processing systems}, 30.

\bibitem[{Madry et~al.(2017)Madry, Makelov, Schmidt, Tsipras, and Vladu}]{madry2017towards}
Madry, A.; Makelov, A.; Schmidt, L.; Tsipras, D.; and Vladu, A. 2017.
\newblock Towards deep learning models resistant to adversarial attacks.
\newblock \emph{arXiv preprint arXiv:1706.06083}.

\bibitem[{Peng et~al.(2021)Peng, Rashid, Schroeder~de Witt, Kamienny, Torr, B{\"o}hmer, and Whiteson}]{peng2021facmac}
Peng, B.; Rashid, T.; Schroeder~de Witt, C.; Kamienny, P.-A.; Torr, P.; B{\"o}hmer, W.; and Whiteson, S. 2021.
\newblock Facmac: Factored multi-agent centralised policy gradients.
\newblock \emph{Advances in Neural Information Processing Systems}, 34: 12208--12221.

\bibitem[{Shapley(1953)}]{shapley1953stochastic}
Shapley, L.~S. 1953.
\newblock Stochastic games.
\newblock \emph{Proceedings of the national academy of sciences}, 39(10): 1095--1100.

\bibitem[{Sun, Kim, and How(2022)}]{sun2022romax}
Sun, C.; Kim, D.-K.; and How, J.~P. 2022.
\newblock ROMAX: certifiably robust deep multiagent reinforcement learning via convex relaxation.
\newblock In \emph{2022 International Conference on Robotics and Automation (ICRA)}, 5503--5510. IEEE.

\bibitem[{Sun et~al.(2021)Sun, Zheng, Liang, and Huang}]{sun2021strongest}
Sun, Y.; Zheng, R.; Liang, Y.; and Huang, F. 2021.
\newblock Who is the strongest enemy? towards optimal and efficient evasion attacks in deep rl.
\newblock \emph{arXiv preprint arXiv:2106.05087}.

\bibitem[{Tessler, Efroni, and Mannor(2019)}]{tessler2019action}
Tessler, C.; Efroni, Y.; and Mannor, S. 2019.
\newblock Action robust reinforcement learning and applications in continuous control.
\newblock In \emph{International Conference on Machine Learning}, 6215--6224. PMLR.

\bibitem[{Tu et~al.(2021)Tu, Wang, Wang, Manivasagam, Ren, and Urtasun}]{tu2021adversarial}
Tu, J.; Wang, T.; Wang, J.; Manivasagam, S.; Ren, M.; and Urtasun, R. 2021.
\newblock Adversarial attacks on multi-agent communication.
\newblock In \emph{Proceedings of the IEEE/CVF International Conference on Computer Vision}, 7768--7777.

\bibitem[{Vinyals et~al.(2019)Vinyals, Babuschkin, Chung, Mathieu, Jaderberg, Czarnecki, Dudzik, Huang, Georgiev, Powell et~al.}]{vinyals2019alphastar}
Vinyals, O.; Babuschkin, I.; Chung, J.; Mathieu, M.; Jaderberg, M.; Czarnecki, W.~M.; Dudzik, A.; Huang, A.; Georgiev, P.; Powell, R.; et~al. 2019.
\newblock Alphastar: Mastering the real-time strategy game starcraft ii.
\newblock \emph{DeepMind blog}, 2: 20.

\bibitem[{Whiteson et~al.(2019)Whiteson, Samvelyan, Rashid, De~Witt, Farquhar, Nardelli, Rudner, Hung, Torr, and Foerster}]{whiteson2019starcraft}
Whiteson, S.; Samvelyan, M.; Rashid, T.; De~Witt, C.; Farquhar, G.; Nardelli, N.; Rudner, T.; Hung, C.; Torr, P.; and Foerster, J. 2019.
\newblock The StarCraft multi-agent challenge.
\newblock In \emph{Proceedings of the International Joint Conference on Autonomous Agents and Multiagent Systems, AAMAS}, 2186--2188.

\bibitem[{Yu et~al.(2022)Yu, Velu, Vinitsky, Gao, Wang, Bayen, and Wu}]{yu2022surprising}
Yu, C.; Velu, A.; Vinitsky, E.; Gao, J.; Wang, Y.; Bayen, A.; and Wu, Y. 2022.
\newblock The surprising effectiveness of ppo in cooperative multi-agent games.
\newblock \emph{Advances in neural information processing systems}, 35: 24611--24624.

\bibitem[{Zhang et~al.(2021)Zhang, Chen, Boning, and Hsieh}]{zhang2021robust}
Zhang, H.; Chen, H.; Boning, D.; and Hsieh, C.-J. 2021.
\newblock Robust reinforcement learning on state observations with learned optimal adversary.
\newblock \emph{arXiv preprint arXiv:2101.08452}.

\bibitem[{Zhang et~al.(2020)Zhang, Chen, Xiao, Li, Liu, Boning, and Hsieh}]{zhang2020robust}
Zhang, H.; Chen, H.; Xiao, C.; Li, B.; Liu, M.; Boning, D.; and Hsieh, C.-J. 2020.
\newblock Robust deep reinforcement learning against adversarial perturbations on state observations.
\newblock \emph{Advances in neural information processing systems}, 33: 21024--21037.

\bibitem[{Zhang et~al.(2022)Zhang, Liu, Xiong, Xu, Wang, Xin, and Wu}]{zhang2022rlcharge}
Zhang, W.; Liu, H.; Xiong, H.; Xu, T.; Wang, F.; Xin, H.; and Wu, H. 2022.
\newblock RLCharge: Imitative multi-agent spatiotemporal reinforcement learning for electric vehicle charging station recommendation.
\newblock \emph{IEEE Transactions on Knowledge and Data Engineering}, 35(6): 6290--6304.

\bibitem[{Zhou and Liu(2023)}]{zhou2023robustness}
Zhou, Z.; and Liu, G. 2023.
\newblock Robustness Testing for Multi-Agent Reinforcement Learning: State Perturbations on Critical Agents.
\newblock In \emph{ECAI}, 3131--3139.

\bibitem[{Zhou et~al.(2024)Zhou, Liu, Guo, and Zhou}]{zhou2024adversarial}
Zhou, Z.; Liu, G.; Guo, W.; and Zhou, M. 2024.
\newblock Adversarial attacks on multiagent deep reinforcement learning models in continuous action space.
\newblock \emph{IEEE Transactions on Systems, Man, and Cybernetics: Systems}.

\end{thebibliography}


\clearpage
\appendix
\section{Proofs of Theorems}
\label{section:appendix_a}

\subsection{Proof of Theorem~\ref{theorem:critic}}
\begin{proof}
Let $V^*(s)$ be the value function under the globally optimal attack, and $V^h(s)$ be the value under our heuristic single-step attack. Let $\mathcal{T}^*$ and $\mathcal{T}^h$ be their respective Bellman operators. $V^*$ and $V^h$ are the fixed points of these operators. The gap is $G = V^h - V^*$.

\begin{align}
    (\mathcal{T}^{*}V)(s) &= \min_{\delta_s, \delta_a} \mathbb{E}_{\mathbf{a}''}[r(s, \mathbf{a}'') + \gamma \mathbb{E}_{s'}[V(s')]] \\
    (\mathcal{T}^{h}V)(s) &= \mathbb{E}_{\mathbf{a}''_{h}}[r(s, \mathbf{a}''_{h}) + \gamma \mathbb{E}_{s'}[V(s')]]
\end{align}

where $\mathbf{a}''_{h}$ results from the heuristic attack minimizing $Q(s+\delta_s, \mu(s+\delta_s)+\delta_a)$.

The gap can be written as:
\begin{equation}
\begin{split}
    G &= \mathcal{T}^h V^h - \mathcal{T}^* V^* \\
    &= (\mathcal{T}^h V^h - \mathcal{T}^* V^h) + (\mathcal{T}^* V^h - \mathcal{T}^* V^*)
\end{split}
\end{equation}

Since $\mathcal{T}^*$ is a $\gamma$-contraction, taking the infinity norm yields:

\begin{equation}
    \|G\|_{\infty} \le \|\mathcal{T}^h V^h - \mathcal{T}^* V^h\|_{\infty} + \gamma \|G\|_{\infty}
\label{eq:gap_recursion_simple}
\end{equation}

The first term is the single-step error. Let
\begin{equation}
    J(\delta_s, \delta_a; V) = \mathbb{E}_{\mathbf{a}''}[r(s, \mathbf{a}'') + \gamma \mathbb{E}_{s'}[V(s')]].
\end{equation} 
Then we have
\begin{equation}
    (\mathcal{T}^*V)(s) = \min_{\delta_s, \delta_a} J(\delta_s, \delta_a; V).
\end{equation}

Let $(\delta_{s,h}, \delta_{a,h})$ be the perturbations found by the heuristic attack. We have

\begin{equation}
\begin{split}
    &(\mathcal{T}^h V^h)(s) - (\mathcal{T}^* V^h)(s) \\
    &= J(\delta_{s,h}, \delta_{a,h}; V^h) - \min_{\delta_s, \delta_a} J(\delta_s, \delta_a; V^h) \\
    &\le J(\delta_{s,h}, \delta_{a,h}; V^\Pi) - \min_{\delta_s, \delta_a} J(\delta_s, \delta_a; V^\Pi) \\
    &+ O(\gamma \|V^\Pi - V^h\|_\infty).
\end{split}
\end{equation}

The heuristic minimizes $Q$, which is an $\epsilon_Q$-approximation of $V^\Pi$. Let $(\delta_s^*, \delta_a^*)$ be the minimizer of $J(\cdot; V^\Pi)$.

\begin{equation}
\begin{split}
    J(\delta_{s,h}, \delta_{a,h}; V^\Pi) &\le Q(s+\delta_{s,h}, \mu(s+\delta_{s,h})+\delta_{a,h}) + \epsilon_Q \\
    &\le Q(s+\delta_s^*, \mu(s+\delta_s^*)+\delta_a^*) + \epsilon_Q \\
    &\le J(\delta_s^*, \delta_a^*; V^\Pi) + 2\epsilon_Q
\end{split}
\end{equation}

Ignoring the higher-order error term for simplicity, the single-step error is bounded by $2\epsilon_Q$. Plugging this into Eq. \eqref{eq:gap_recursion_simple}:
\begin{equation}
\begin{split}
    & \|G\|_{\infty} \le 2\epsilon_Q + \gamma \|G\|_{\infty} \\
    & \implies \|V^h - V^*\|_{\infty} \le \frac{2\epsilon_Q}{1-\gamma}
\end{split}
\end{equation}

This completes the proof.
\end{proof}

\subsection{Proof of Theorem~\ref{theorem:bounded_reward_degradation}}
\begin{proof}
The value functions are defined by the Bellman equations:
\begin{align}
    V^{\Pi}(s) &= \mathbb{E}_{a \sim \Pi(\cdot|s)} [r(s,a) + \gamma \mathbb{E}_{s' \sim P(\cdot|s,a)}[V^{\Pi}(s')]] \\
    V^{\Pi}_{\text{adv}}(s) &= \mathbb{E}_{a \sim \Pi(\cdot|\hat{s})} [r(s,a) + \gamma \mathbb{E}_{s' \sim P(\cdot|s,a)}[V^{\Pi}_{\text{adv}}(s')]]
\end{align}

Let $\delta(s) = V^{\Pi}(s) - V^{\Pi}_{\text{adv}}(s) $.
\begin{equation}
\begin{split}
    \delta(s) &= \mathbb{E}_{a \sim \Pi(\cdot|s)} [r(s,a)] - \mathbb{E}_{a \sim \Pi(\cdot|\hat{s})} [r(s,a)] \\
    &+ \gamma \left( \mathbb{E}_{a \sim \Pi(\cdot|s)} [V^{\Pi}(s')] - \mathbb{E}_{a \sim \Pi(\cdot|\hat{s})} [V^{\Pi}_{\text{adv}}(s')] \right) 
\end{split}
\label{eq:delta_expand}
\end{equation}

The first term is the expected reward difference due to the change in the action distribution.

\begin{equation}
\begin{split}
    \left| \mathbb{E}_{a \sim \Pi(\cdot|s)} [r(s,a)] - \mathbb{E}_{a \sim \Pi(\cdot|\hat{s})} [r(s,a)] \right| \\
    \le 2 R_{\max} \cdot D_{\text{TV}}(\Pi(\cdot|s) \,||\, \Pi(\cdot|\hat{s})) 
\end{split}    
\label{eq:reward_bound}
\end{equation}

where $R_{\max} = \sup_{s,a} |r(s,a)|$ is the maximum possible absolute reward, and $D_{\text{TV}}$ is the total variation distance.

For the second term of Eq. \eqref{eq:delta_expand}, We use the triangle inequality:

\begin{equation}
\begin{split}    
    \left| \mathbb{E}_{a \sim \Pi(\cdot|s)} [V^{\Pi}(s')] - \mathbb{E}_{a \sim \Pi(\cdot|\hat{s})} [V^{\Pi}_{\text{adv}}(s')] \right| \\
    \le \left| \mathbb{E}_{a \sim \Pi(\cdot|s)} [V^{\Pi}(s')] - \mathbb{E}_{a \sim \Pi(\cdot|\hat{s})} [V^{\Pi}(s')] \right| \\
    \quad + \left| \mathbb{E}_{a \sim \Pi(\cdot|\hat{s})} [V^{\Pi}(s')] - \mathbb{E}_{a \sim \Pi(\cdot|\hat{s})} [V^{\Pi}_{\text{adv}}(s')] \right| 
\end{split}   
\label{eq:value_split}
\end{equation}

The first part of Eq. \eqref{eq:value_split} is bounded similarly to the reward term:

\begin{equation}
\begin{split}
    \left| \mathbb{E}_{a \sim \Pi(\cdot|s)} [V^{\Pi}(s')] - \mathbb{E}_{a \sim \Pi(\cdot|\hat{s})} [V^{\Pi}(s')] \right| \\
    \le 2 V_{\max} \cdot D_{\text{TV}}(\Pi(\cdot|s) \,||\, \Pi(\cdot|\hat{s}))
\end{split}
\end{equation}

where $V_{\max} = R_{\max} / (1-\gamma)$. The second part of Eq. \eqref{eq:value_split} can be bounded as:

\begin{equation}
\begin{split}
    &\left| \mathbb{E}_{a \sim \Pi(\cdot|\hat{s})} [V^{\Pi}(s') - V^{\Pi}_{\text{adv}}(s')] \right| \\
    &= \left| \mathbb{E}_{a \sim \Pi(\cdot|\hat{s})} [\delta(s')] \right| \\
    &\le \mathbb{E}_{a \sim \Pi(\cdot|\hat{s})} [|\delta(s')|] \\
    &\le \max_{s'} |\delta(s')|
\end{split}
\end{equation}

The joint policy $\Pi(\cdot|s)$ is a product of independent individual policies $\pi_i(\cdot|s)$. For two such product distributions, the total variation distance is bounded by the sum of the total variation distances of the marginals:

\begin{equation}
    D_{\text{TV}}(\Pi(\cdot|s) \,||\, \Pi(\cdot|\hat{s})) \le \sum_{i=1}^{n} D_{\text{TV}}(\pi_i(\cdot|s) \,||\, \pi_i(\cdot|\hat{s}))
\end{equation}

By applying Pinsker's inequality, $D_{\text{TV}}(P||Q) \le \sqrt{\frac{1}{2}D_{\text{KL}}(P||Q)}$, to each term:

\begin{equation}
\begin{split}
    \sum_{i=1}^{n} D_{\text{TV}}(\pi_i(\cdot|s) \,||\, \pi_i(\cdot|\hat{s})) \\
    \le \sum_{i=1}^{n} \sqrt{\frac{1}{2}D_{\text{KL}}(\pi_i(\cdot|s) \,||\, \pi_i(\cdot|\hat{s}))}
\end{split}
\end{equation}

Using the inequality for sums of square roots $(\sum \sqrt{x_i} \le \sqrt{n \sum x_i})$, which follows from Cauchy-Schwarz:

\begin{equation}
\begin{split}
    &\sum_{i=1}^{n} \sqrt{\frac{1}{2}D_{\text{KL}}(\pi_i(\cdot|s) \,||\, \pi_i(\cdot|\hat{s}))} \\
    &\le \sqrt{n \sum_{i=1}^{n} \frac{1}{2}D_{\text{KL}}(\pi_i(\cdot|s) \,||\, \pi_i(\cdot|\hat{s}))} \\
    &= \sqrt{\frac{n}{2} \mathcal{D}_{\Pi}(s, \hat{s})}
\end{split}
\label{eq:tv_to_kl_sum}
\end{equation}

where $\mathcal{D}_{\Pi}(s, \hat{s})$ is the Multi-Agent Action Difference as defined in Eq. \eqref{eq:ma_ad}.

Let $\epsilon_s = \max_{\hat{s} \in \mathcal{B}(s)} \sqrt{\mathcal{D}_{\Pi}(s, \hat{s})}$.
Combining the bounds, we have:

\begin{equation}
\begin{split}
    &|\delta(s)| \le 2 R_{\max} D_{\text{TV}}(\Pi(\cdot|s) \,||\, \Pi(\cdot|\hat{s})) \\
    &+\gamma \left( 2 V_{\max} D_{\text{TV}}(\Pi(\cdot|s) \,||\, \Pi(\cdot|\hat{s})) + \max_{s'} |\delta(s')| \right)  \\
    &\le 2 (R_{\max} + \gamma V_{\max}) \sqrt{\frac{n}{2}} \cdot \max_{\hat{s} \in \mathcal{B}(s)} \sqrt{\mathcal{D}_{\Pi}(s, \hat{s})} + \gamma \max_{s'} |\delta(s')|  \\
    &\le \frac{2 R_{\max}}{1-\gamma} \sqrt{\frac{n}{2}} \cdot \epsilon_s + \gamma \max_{s'} |\delta(s')|
\end{split}
\end{equation}

Let $\epsilon = \sup_s |\delta(s)|$. Since the inequality holds for any $s$, it must also hold for the supremum:

\begin{equation}
    \epsilon \le \frac{2 R_{\max}}{1-\gamma} \sqrt{\frac{n}{2}} \cdot \epsilon_s + \gamma \epsilon
\end{equation}

Solving for $\epsilon$:

\begin{equation}
\begin{split}
    (1-\gamma)\epsilon &\le \frac{2 R_{\max}}{1-\gamma} \sqrt{\frac{n}{2}} \cdot \epsilon_s \\
    \implies \epsilon &\le \frac{2 R_{\max}}{(1-\gamma)^2} \sqrt{\frac{n}{2}} \cdot \epsilon_s
\end{split}
\end{equation}

By defining the constant $C = \frac{\sqrt{2n} R_{\max}}{(1-\gamma)^2}$, we arrive at the final result:

\begin{equation}
    |V^{\Pi}(s) - V^{\Pi}_{\text{adv}}(s)| \le C \cdot \max_{\hat{s} \in \mathcal{B}(s)} \sqrt{\mathcal{D}_{\Pi}(s, \hat{s})}
\end{equation}

This completes the proof.
\end{proof}

\section{Details of Defense Mechanisms}
\label{section:appendix_b}
We adopted three adversarial defense (PAAD,ATLA,M3DDPG) in our experiments to evaluate the effectiveness of SAJA. This section details our implementation of these methods.

\subsection{PAAD}
PAAD (Policy Adversarial Actor-Director) is a two-step state defense method. In our experiments, the method trains a separate "director" network that learns to guide the system toward low-reward states. At each step of adversarial training: The director network generates a worst-case target action $a_{\text{tar}}$ from the current state $s_t$. A PGD-based optimization then finds a state perturbation $\delta_s$ that forces the main policy to produce an action that closely matches $a_{\text{tar}}$ when given the perturbed state $s_t + \delta_s$. This resulting adversarial state $s_t + \delta_s$ is then used to train the main policy for enhanced robustness.

\subsection{ATLA}
ATLA (Alternating Training with Learned Adversary) is a state defense that improves robustness by training the main policy against a learned adversary. In our implementation, this adversary is a separate MAPPO (Multi-Agent PPO) agent that takes the clean observation $s_t$ as input and outputs a perturbation vector $\delta_s$ as its action. The adversary is trained to maximize the negative team reward ($-r_t$) of the main policy, thereby learning to generate worst-case perturbations. During training, the main policy receives the adversarially perturbed observation $s'_t = s_t + \delta_s$ and is updated with the true environmental reward $r_t$. This co-evolutionary process forces the main policy to become robust against a dynamically adapting threat. Key hyperparameters for the MAPPO adversary include: a hidden dimension of 64, $\gamma=0.98$, $\lambda=0.9$, and a PPO clipping $\varepsilon=0.2$.

\subsection{M3DDPG}
M3DDPG (Minimax Multi-Agent DDPG) (Li et al. 2019) is an action defense method that improves robustness by training agents against worst-case adversarial actions from their counterparts. The original algorithm assumes an individual critic for each agent. Our framework uses a common architectural variant with a single, centralized critic that outputs individual Q-values for each agent ($Q(s, a) \rightarrow [Q_1, \dots, Q_N]$). To adapt M3DDPG to this shared-critic setup, we made the following key modifications, detailed in Algorithm~\ref{alg:m3ddpg_adapted}:

\paragraph{Critic Update} Instead of perturbing each agent's target action individually, we apply a single adversarial perturbation $\delta_a$ to the joint target action $a'$. This perturbation is derived from the gradient of the target critic $Q'$ and scaled by a hyperparameter $\varepsilon_{\text{adv}}$ (set to $0.001$).

\paragraph{Actor Update} We follow the original minimax principle. To update agent $i$'s actor, we first compute an adversarial joint action $a_{\text{adv}}$ using the current critic $Q$. We then calculate the policy gradient for agent $i$ using a modified joint action where the $i$-th component of $a_{\text{adv}}$ is replaced with the new action from actor $\mu_i$.
This approach preserves the core minimax spirit of M3DDPG while making it compatible with modern centralized critic architectures.

\begin{algorithm}[h!]
\caption{Our Implementation of M3DDPG}
\label{alg:m3ddpg_adapted}
\begin{algorithmic}[1] 
\FOR{episode = 1 \TO $M$}
    \STATE Initialize exploration noise $\mathcal{N}$, receive initial state $s$.
    \FOR{$t = 1$ \TO max-episode-length}
        \STATE For each agent $i$, select action $a_i = \mu_i(o_i) + \mathcal{N}_t$.
        \STATE Execute joint action $a = (a_1, \dots, a_N)$, observe reward $r$ and new state $s'$.
        \STATE Store $(s, a, r, s')$ in replay buffer $\mathcal{D}$.
    \ENDFOR
    \STATE Sample a random minibatch of $S$ transitions $(s^k, a^k, r^k, s'^k)$ from $\mathcal{D}$.
    
    \STATE \textit{// --- Critic Update ---}
    \STATE Get joint target action: $a'^k = \mu'(s'^k)$.
    \STATE Compute perturbation for joint target action: $\delta_{a'} \leftarrow -\varepsilon_{\text{adv}} \cdot \nabla_{a'^k} Q'(s'^k, a'^k)$.
    \STATE Compute TD target for all agents using the perturbed joint action:
    \STATE $y^k \leftarrow r^k + \gamma Q'(s'^k, a'^k + \delta_{a'})$.
    \STATE Update shared critic by minimizing the loss $\mathcal{L}(\phi) = \frac{1}{S} \sum_k (y^k - Q(s^k, a^k))^2$.
    
    \STATE \textit{// --- Actor Update ---}
    \STATE Compute perturbation for current joint action: $\delta_a \leftarrow -\varepsilon_{\text{adv}} \cdot \nabla_{a^k} Q(s^k, a^k)$.
    \STATE $a_{\text{adv}}^k \leftarrow a^k + \delta_a$.
    \FOR{agent $i=1$ \TO $N$}
        \STATE Construct action for policy gradient: $\tilde{a}^k = (a_{\text{adv}, 1}^k, \dots, \mu_i(o_i^k), \dots, a_{\text{adv}, N}^k)$.
        \STATE Update actor $\mu_i$ using the sampled policy gradient:
        \STATE $\nabla_{\theta_i} J \approx \frac{1}{S} \sum_k \nabla_{\theta_i} \mu_i(o_i^k) \nabla_{a_i} Q_i(s^k, \tilde{a}^k) |_{a_i=\mu_i(o_i^k)}$.
    \ENDFOR
    
    \STATE Update target networks for all agents.
\ENDFOR
\end{algorithmic}
\end{algorithm}

\section{Details of Benchmark Attack Methods}
\label{section:appendix_c}
Below, we detail the five benchmark methods used in our evaluation. They are categorized into two classes: Stochastic and Gradient-based attacks.

\subsection{Stochastic Methods}
This class of baselines is designed to evaluate whether SAJA's effectiveness stems from its two-phase architecture, gradient-based optimization, or merely from the introduction of noise.

\begin{itemize}
    \item \textbf{Random-State:} For each victim agent's observation vector $s$, we first generate a random vector of the same dimension, with each element sampled from a uniform distribution $U(-1, 1)$. We then take the sign of this random vector and scale it by the state perturbation budget $\epsilon_s$ to create the final noise $\delta_s \leftarrow \varepsilon_s \cdot \text{sign}(r_s)$. This noise is added to the original observation $s_V^* \leftarrow s_V + \delta_{s, V}$. The states of the other $n-m$ non-victim agents remain unperturbed $s_{\neg V}^* \leftarrow s_{\neg V}$. 

    \item \textbf{Random-Action:} After all agents generate actions $a$ based on its  unperturbed state $s$, we follow the exact same procedure as in Random-State to generate random action noise, but using the action perturbation budget $\epsilon_a$. This noise is added to the action $a_V^{*} \leftarrow a_V + \delta_{a, V}$. The actions of the other $n-m$ non-victim agents remain unperturbed $a_{\neg V}^{*} \leftarrow a_{\neg V}$. 

    \item \textbf{Random-State-Action:} This method mimics the two-phase structure of SAJA but replaces the gradient-based attacks with random perturbations. It first performs a Random-State attack to generate a perturbed state $s^*$. Then, $s^*$ is passed through the policy network to obtain an intermediate action $a'$. Finally, it applies a Random-Action attack to this intermediate action $a'$ to yield the final executed action $a^{**}$. The detailed procedure is outlined in Algorithm~\ref{alg:rsa}. Notably, setting the state perturbation budget $\epsilon_s=0$ recovers the Random-Action attack, while setting the action perturbation budget $\epsilon_a=0$ recovers the Random-State attack.
\end{itemize}

\begin{algorithm}[h!]
\caption{Random-State-Action Attack}
\label{alg:rsa}
\textbf{Input}: Original state $s$, actor networks $\mu$, perturbation budgets $\varepsilon_s, \varepsilon_a$. \\
\textbf{Output}: Adversarial state-action pair $(s^*, a^{**})$. \\
\begin{algorithmic}[1] 
\STATE Get original state $s$.
\STATE Randomly select a set $V$ of $m$ victims from $\{1, \dots, n\}$.

\STATE \textit{// --- Random State Attack Phase ---}
\STATE Generate random vector $r_s \sim U(-1, 1)$ with the same shape as $s$.
\STATE Compute state perturbation $\delta_s \leftarrow \varepsilon_s \cdot \text{sign}(r_s)$.
\STATE Apply perturbation to victims' states: $s_V^* \leftarrow s_V + \delta_{s, V}$.
\STATE Ensure all other states remain unchanged: $s_{\neg V}^* \leftarrow s_{\neg V}$.

\STATE \textit{// --- Random Action Attack Phase ---}
\STATE Compute intermediate action from perturbed state: $a' \leftarrow \mu(s^*)$.
\STATE Generate random vector $r_a \sim U(-1, 1)$ with the same shape as $a'$.
\STATE Compute action perturbation $\delta_a \leftarrow \varepsilon_a \cdot \text{sign}(r_a)$.
\STATE Apply perturbation to victims' actions: $a_V^{**} \leftarrow a'_V + \delta_{a, V}$.
\STATE Ensure all other actions remain unchanged: $a_{\neg V}^{**} \leftarrow a'_{\neg V}$.

\STATE \textbf{return} $(s^*, a^{**})$
\end{algorithmic}
\end{algorithm}

\subsection{Gradient-based Ablation Baselines}
Our gradient-based baselines are designed as direct ablations of the SAJA framework:

\begin{itemize}
    \item \textbf{PGD-State:} This method focuses exclusively on state perturbation. It is equivalent to running the full SAJA algorithm (Algorithm~\ref{alg:saja}) with the number of action attack iterations set to zero, i.e., $K_a=0$.
    
    \item \textbf{PGD-Action:} This method focuses exclusively on action perturbation. It is equivalent to running the full SAJA algorithm (Algorithm~\ref{alg:saja}) with the number of state attack iterations set to zero, i.e., $K_s=0$.
\end{itemize}

This formulation makes it clear that PGD-State and PGD-Action are not merely conceptually related to SAJA, but are true special cases of our framework, allowing for a rigorous and fair ablation study.

\section{Detailed Experimental Setup}
\label{section:appendix_d}
This section provides the detailed configurations required to reproduce our experimental results, including the hyperparameters for model training, the neural network architectures, and the base model training procedures.

\subsection{MADDPG}
Our implementation of MADDPG uses the Adam optimizer for both the actor and critic networks. The learning rate for the actors was set to $0.01$, while the critic learning rate was set to $0.05$. We used a discount factor $\gamma$ of $0.85$ and a TD-Lambda ($\lambda$) of $0.8$. The models were trained with a batch size of $32$, sampling from a replay buffer with a capacity of $5000$ experiences. For exploration, Gaussian noise with a standard deviation of $0.1$ was added to the actor's actions during training. Target networks were updated using a soft update scheme with a rate $\tau = 0.001$. For regularization, we applied a gradient norm clip of $0.5$ and a weight decay factor of $1 \times 10^{-4}$ to the critic optimizer. When training the M3DDPG defense, the adversarial perturbation budget $\epsilon$ for actions was set to $0.001$.

The actor and critic networks are both based on Multi-Layer Perceptrons (MLPs). Each agent's actor network is a Recurrent Neural Network (RNN) that processes local observations. It consists of an input fully connected layer, a Gated Recurrent Unit (GRU) cell, and an output fully connected layer, all with a hidden dimension of $64$. A $\tanh$ activation is applied to the final output to produce actions in the range $[-1, 1]$. The centralized critic takes the global state and the joint action of all agents as input. It is a feed-forward network with two hidden layers of size $64$ using ReLU activations, followed by a linear layer that outputs a single Q-value.

\subsection{FACMAC}
All FACMAC models were trained using the Adam optimizer. The learning rates were set to $0.05$ for the actors and $0.01$ for the critic and mixer networks, with an optimizer epsilon of $0.05$. We used a discount factor $\gamma$ of $0.85$ and a soft target update coefficient $\tau$ of $0.001$. Training was performed with a batch size of $32$ from a replay buffer of size $5000$. Gaussian noise with a standard deviation of $0.1$ was used for exploration. The QMixer was configured with an embedding dimension of $64$.

The actor network for each agent is identical to the recurrent GRU-based architecture used in our MADDPG implementation. FACMAC employs individual critic networks for each agent, which take the agent's local observation and action as input. Each critic is a 3-layer MLP (64 hidden units, ReLU activations) outputting a single Q-value. These individual Q-values are then processed by a centralized QMixer. The QMixer uses a state-conditioned hypernetwork to generate weights for a mixing network that non-linearly combines the individual Q-values into a global, joint-action Q-value, $Q_{tot}$, enabling factored value decomposition.

\subsection{Benchmark Model Training}
To ensure the validity and fairness of our attack evaluations, all attacks were conducted on fully trained benchmark models. For the vanilla benchmark models, each MADDPG and FACMAC variant was trained for a total of 2 million timesteps. During training, we evaluated the model's average reward and standard deviation over 10 evaluation episodes every 2000 timesteps, selecting the final model at the 2-million-timestep mark as the attack target. The defended models trained with PAAD and ATLA were trained for 4 million timesteps, while the M3DDPG model was trained for 2 million timesteps. 

\begin{figure}[h!]
    \centering
    \includegraphics[width=\columnwidth]{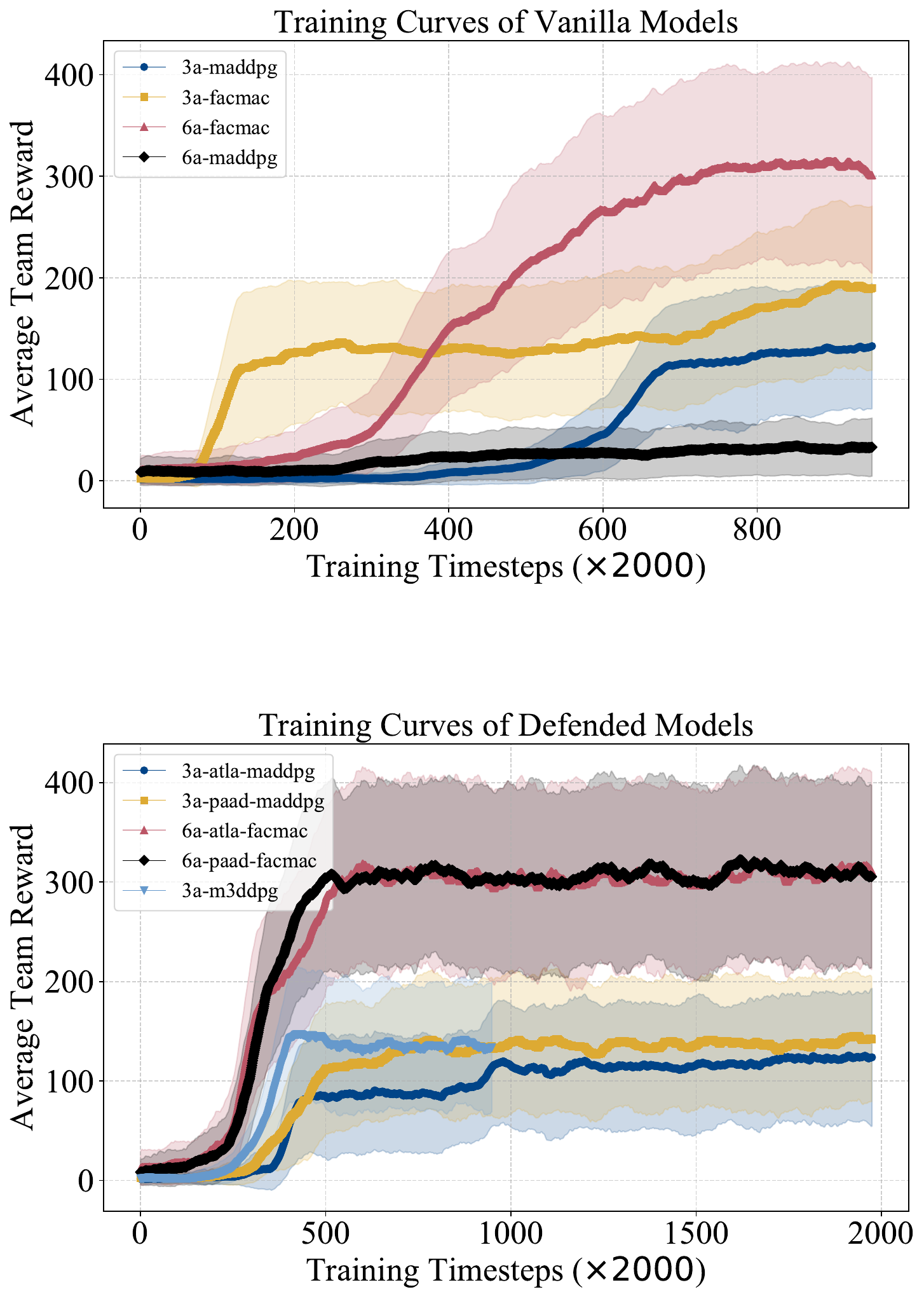}
    \caption{\textbf{(Left)} Learning curves for the vanilla models. \textbf{(Right)} Learning curves for the defended models. Training curves for all benchmark models used in the experiments. The solid lines represent the average team reward over 10 evaluation episodes, and the shaded areas represent the standard deviation.Training curves for all benchmark models used in the experiments.}
    \label{fig:training_curves}
\end{figure}

Figure \ref{fig:training_curves} displays the learning curves for all vanilla and defended models used in our experiments. As shown, all models reached a stable level of performance, providing a solid foundation for our adversarial attack evaluations.

\end{document}